\documentclass[lettersize,journal]{IEEEtran}
\usepackage[table]{xcolor}
\usepackage{amsmath,amsfonts}
\usepackage{algorithmic}
\usepackage{algorithm}
\usepackage{array}
\usepackage[caption=false,font=normalsize,labelfont=sf,textfont=sf]{subfig}
\usepackage{textcomp}
\usepackage{stfloats}
\usepackage{url}
\usepackage{verbatim}
\usepackage{graphicx}
\usepackage{cite}
\usepackage{tikz}
\usepackage{makecell}
\usepackage{multirow}
\usepackage{booktabs}
\usepackage{threeparttable}
\usepackage{color}
\newcommand{\etal}{{et al. }}
\usepackage{hyperref}

\usepackage{orcidlink}
\hypersetup{hidelinks}

\begin{document}


\title{Finger Pose Estimation for Under-screen Fingerprint Sensor}

\author{Xiongjun Guan$^{\orcidlink{0000-0001-8887-3735}}$, 
	Zhiyu Pan$^{\orcidlink{0009-0000-6721-4482}}$,
	Jianjiang Feng$^{\orcidlink{0000-0003-4971-6707}}$, ~\IEEEmembership{Member, IEEE}, 
	and Jie Zhou$^{\orcidlink{0000-0001-7701-234X}}$, ~\IEEEmembership{Fellow, IEEE}
	
	\thanks{
		This work was supported in part by the National Natural Science Foundation of China under Grant 62376132 and 62321005. (\emph{Corresponding author: Jianjiang Feng}.)}
	\IEEEcompsocitemizethanks{
    \IEEEcompsocthanksitem
	The authors are with Department of Automation, Tsinghua University, Beijing 100084, China (e-mail: \url{gxj21@mails.tsinghua.edu.cn}; \url{pzy20@mails.tsinghua.edu.cn}; \url{jfeng@tsinghua.edu.cn}; \url{jzhou@tsinghua.edu.cn}).
	}
	
}

\markboth{IEEE TRANSACTIONS ON INFORMATION FORENSICS AND SECURITY, VOL.~x, MONTH~2025}%
{Shell \MakeLowercase{\textit{et al.}}: A Sample Article Using IEEEtran.cls for IEEE Journals}


\maketitle

\begin{abstract}
Two-dimensional pose estimation plays a crucial role in fingerprint recognition by facilitating global alignment and reduce pose-induced variations.
However, existing methods are still unsatisfactory when handling with large angle or small area inputs.
These limitations are particularly pronounced on fingerprints captured by under-screen fingerprint sensors in smartphones.
In this paper, we present a novel dual-modal input based network for under-screen fingerprint pose estimation. 
Our approach effectively integrates two distinct yet complementary modalities: texture details extracted from ridge patches through the under-screen fingerprint sensor, and rough contours derived from capacitive images obtained via the touch screen. 
This collaborative integration endows our network with more comprehensive and discriminative information, substantially improving the accuracy and stability of pose estimation.
A decoupled probability distribution prediction task is designed, instead of the traditional supervised forms of numerical regression or heatmap voting, to facilitate the training process.
Additionally, we incorporate a Mixture of Experts (MoE) based feature fusion mechanism and a relationship driven cross-domain knowledge transfer strategy to further strengthen feature extraction and fusion capabilities.
Extensive experiments are conducted on several public datasets and two private datasets. 
The results indicate that our method is significantly superior to previous state-of-the-art (SOTA) methods and remarkably boosts the recognition ability of fingerprint recognition algorithms.
Our code is available at https://github.com/XiongjunGuan/DRACO.

\end{abstract}


\section{Introduction}
Two-dimensional pose estimation has been extensively researched in the field of fingerprint \cite{jain2000filterbank,kenneth2003localization,cappelli2009spatial,yoon2013LFIQ,yang2014localized,ouyang2017fingerprint,deerada2020progressive,engelsma2021learning,yin2021joint,duan2023estimating}.
This task aims to determine the fingerprint's center position and rotation direction from an input image, enabling the effective alignment of heterogeneous data within a unified coordinate system \cite{maltoni2022handbook}.
Functioning as a robust global prior, fingerprint pose plays a pivotal role in fingerprint recognition systems, and is typically employed as an essential  preprocessing stage \cite{maltoni2022handbook}.
For example, pose normalization can substantially reduce intra-class differences caused by varying geometric positions, thereby effectively enhancing the generalizability and discriminability of feature extraction \cite{cao2019automated,engelsma2021learning,guan2023regiression,grosz2024AFRNet,pan2024fixed}.
Besides, incorporating supplementary constraints on pose relationships inherently filters out imposter pairs with erroneous spatial correspondences while streamlining the search space, which can significantly improve the accuracy and efficiency of matching algorithms \cite{cappelli2010MCC,su2016fingerprint,cao2017fingerprint,gu2022latent,duan2023estimating,grosz2023latent}.

\begin{figure}[!t]
    \centering
    \includegraphics[width=0.85\linewidth]{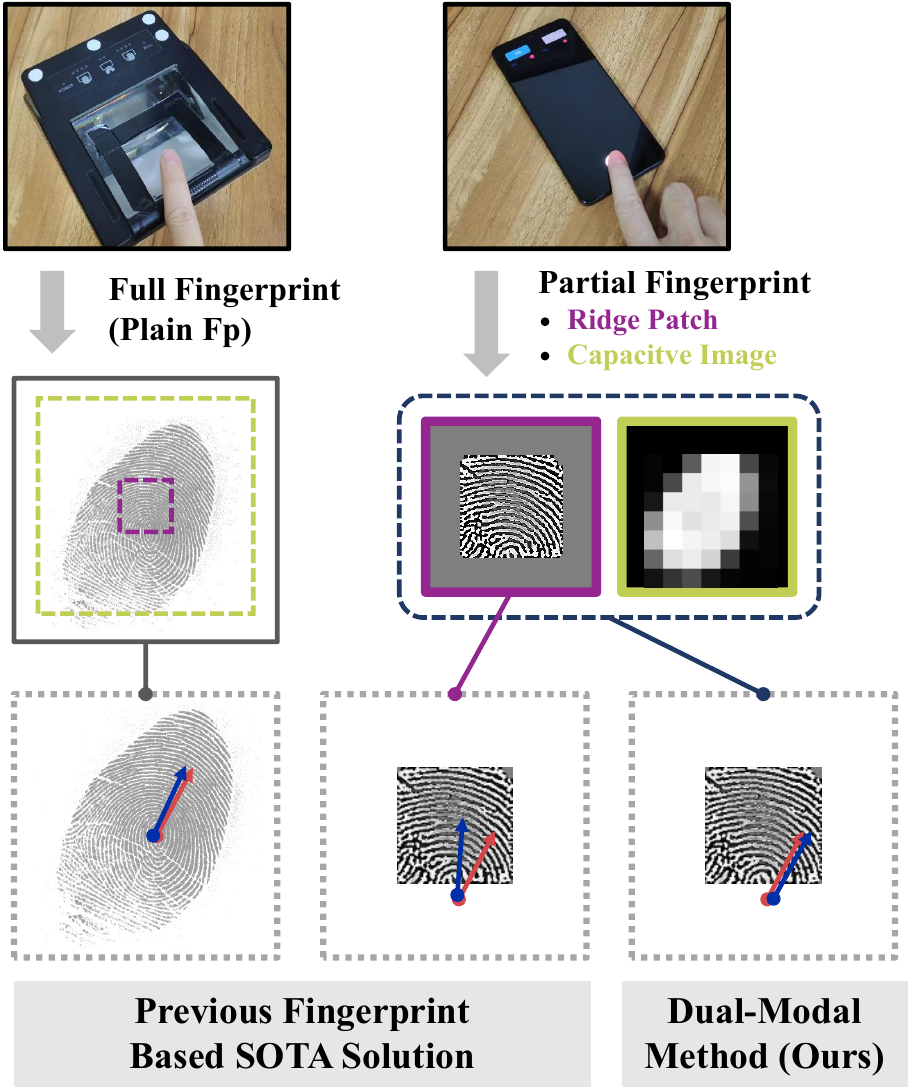}
    \caption{Examples of fingerprint pose estimation under different input modalities. 
    Among them, full fingerprint (plain fingerprint) is collected by a conventional optical fingerprint scanner, while the ridge patch and capacitive image are simultaneously collected from the under screen fingerprint sensor and touch screen of a smartphone (referred to as partial fingerprint collectively) .
    All modalities, captured from the same finger with similar touch gestures, are marked in gray, purple, and green. 
    For clarity, the dashed lines in the full fingerprint indicate the equivalent collection areas for the partial fingerprints.
    Subfigures in the last row represent the estimated result (blue) and ground truth (red) using corresponding modals.
    It can be observed that the performance of previous fingerprint based SOTA solution \cite{duan2023estimating} declines significantly as the available area diminishes,  while our dual-modal method achieves more accurate prediction.
    }
    \label{fig:intro}
\end{figure}

Conventional approaches typically depend on special points \cite{kenneth2003localization,yoon2013LFIQ,deerada2020progressive} or specific areas \cite{jain2000filterbank,yang2014localized,su2016fingerprint,yin2018orientation} to ascertain fingerprint pose. 
Motivated by the success of deep learning, researchers have gradually developed solutions based on neural network in recent years \cite{ouyang2017fingerprint,gu2018efficient,yin2021joint,engelsma2021learning,grosz2023latent,grosz2024AFRNet,duan2023estimating}.
However, these methods are initially designed for rolled or plain fingerprints, which generally necessitate a sufficiently large effective area (about $512 \times 512 \:\text{px}$, $500\:\text{ppi}$) and small angle differences (usually less than $30^\circ$) to acquire adequate information for reliable pose estimation.
In addition to fingerprints, some studies proposed predicting the three-dimensional angle of fingers from capacitive images \cite{xiao2015estimating,mayer2017estimating,he2024TrackPose}, which has been proven effective when inputted at relatively small touch angles (usually within $45^\circ$).
It is imperative to highlight that the mobile device recognition scenario unequivocally surpasses these constraints.
Specifically, with existing under-screen fingerprint sensors, the size of captured image is substantially reduced (about $132 \times 132 \:\text{px}$, $500\:\text{ppi}$), while users may attempt to unlock their devices from arbitrary touch poses, further exacerbating the challenges.

Fig. \ref{fig:intro} shows examples of fingerprints in these mentioned modalities.
To ensure terminological consistency, this paper adopts the following conventions:
(1) \emph{Full Fingerprint} refers to plain fingerprint, distinguishing it from those collected by mobile phones with limited size or resolution,
(2) \emph{Plain Fingerprint}, \emph{Ridge Patch} and \emph{Capacitive Image} denote the specific image types employed in distinct processing streams, and
(3) \emph{Partial Fingerprint} collectively describes the combined data in our experiments that includes the two modalities from smartphones.
It can be intuitively seen that ridge patch and capacitive image exhibit a substantial degradation in terms of available information compared to plain fingerprint.
Moreover, these two modalities from mobile devices exhibit significant differences:
high resolution ridge patches contain rich local structures but limited receptive field, while low resolution capacitive images roughly represent the global contour but lacking localized details.
This exciting complementarity motivates us to explore the enormous potential of modal fusion implementation.
What's more, the widespread adoption of mobile devices equipped with under screen fingerprint sensors and touch screens has created an ideal ecosystem for applying this dual modal approach. 
These devices inherently support the simultaneous acquisition of ridge patches and capacitive images, making the proposed fusion method highly practical. 
Furthermore, this technology can be seamlessly integrated into existing devices through software updates, ensuring broad applicability across various scenarios.
Overall, this innovative dual modal paradigm has convincing development value.

In this paper, we introduce a partial fingerprint pose estimation framework that effectively exploits such complementary strengths.
The proposed approach leverages the collaborative potential of \textbf{D}ual-modal guidance from \textbf{R}idge patches \textbf{A}nd \textbf{C}apacitive images to \textbf{O}ptimize the feature extraction, fusion and representation, called \textbf{DRACO}.
Different from the previous supervision forms of numerical regression \cite{ouyang2017fingerprint,yin2021joint,engelsma2021learning,grosz2023latent,grosz2024AFRNet} or heatmap voting \cite{gu2018efficient,duan2023estimating}, we transform pose representations into decoupled quantized probability distribution, inspired by \cite{li2021TokenPose,li2021human,li2022SimCC,jiang2023Rtmose}.
This upgrade enables our network to better grasp the relative relationships between adjacent pose spaces, resulting in impressive performance gains.
We also apply the MoE mechanism to improve the feature fusion stage.
A lightweight router is employed to dynamically generate adaptive weights for multiple separated feature branches, ensuring an appropriate balance of significance across different modal information.
Furthermore, we leveraged the comparative relationships between groups to facilitate knowledge transfer from the high-performance plain fingerprint pose estimation domain to the target partial fingerprint domain, further strengthening the feature extraction part.

Extensive experiments were conducted on two public fingerprint databases and two private databases.
The results strongly demonstrate the effectiveness of integrated strategies and mechanisms in DRACO.
In addition, the proposed algorithm significantly outperforms existing SOTA pose estimation algorithms in terms of accuracy and stability.
Moreover, we also evaluated the assistance of incorporating pose information for fingerprint recognition, where our approach demonstrated consistent leading performance.

The main contributions of this work can be summarized as:
\begin{itemize}
	\item We propose DRACO, a dual modal partial fingerprint pose estimation framework. 
    The novel multimodality of ridge patch and capacitive image is explored and demonstrated to exhibit significant complementary advantages.
    \item Several simple but effective strategies and mechanisms are introduced to improve the feature extraction, fusion, and representation stages in pose estimation networks, including knowledge transfer, MoE, and decoupled probability distribution. 
    We believe that these evolutions have substantial reference value and may provide potential inspiration for following studies.
    \item Extensive experiments were conducted to comprehensively evaluate the performance of DRACO and existing SOTA methods.
    The experimental results strongly demonstrated the superiority of our proposed approach in terms of both precision and robustness.

\end{itemize}

\section{Related Work}
In this section, we first introduce the definition of fingerprint pose, and then review relevant finger pose estimation algorithms based on fingerprints (plain fingerprints or ridge patches) and capacitive images.

\subsection{Definition of Fingerprint Pose}
Owing to the absence of adequately distinct and consistent anatomical landmarks, the scientific community has yet to establish a clear and unified definition of fingerprint pose \cite{mangold2016data,maltoni2022handbook}.
Researchers have proposed multiple approaches to describe the center and direction of fingerprints in order to achieve approximate goals.
Early studies employed the centroid and positive direction of foreground mask for pose estimation \cite{candela1995pcasys,cappelli2009spatial,watson2009slapssegii}.
In addition, some approaches determined pose parameters through singular points \cite{bazen2001segmentation,kenneth2003localization}.
However, these approaches demonstrate unsatisfactory practicality, as their accuracy substantially depends on the completeness and quality of acquired fingerprint areas.
To address these challenges, subsequent researchers proposed various fingerprint pose definitions based on special patterns along ridge orientation fields, such as points of maximum curvature \cite{liu2005fingerprint,yoon2013LFIQ}, points that match the reference templates \cite{candela1995pcasys,jain2000filterbank}, or focal points perpendicular to the ridges \cite{rerkrai2000new,areekul2006new,deerada2020progressive}.
Despite the improvement in accuracy, fingerprint centers under these definitions still cannot guarantee sufficient consistency for different impressions.

Further integrating these features, Yang \etal \cite{yang2014localized} defined the fingerprint direction as perpendicular to the ridge orientation around the knuckle region,  and determined the center based on the type and number of singular points.
On this basis, Si \etal \cite{si2015detection} introduced a refined approach that utilizes solely the central singular point located at the northernmost as the fingerprint center, while maintaining the same directional definition.
In cases where such a singular point is absent, the point exhibiting the highest curvature is designated as the center.
Subsequent researches \cite{su2016fingerprint,ouyang2017fingerprint,gu2018efficient,yin2021joint,duan2023estimating} followed this form of definition.
In the paper, we also adopt the same definition for consistency and comparability.

\subsection{Pose Estimation Based on Fingerprint}
Traditional methods typically estimate pose information through foreground mask information \cite{candela1995pcasys,cappelli2009spatial,watson2009slapssegii} or detecting special points \cite{bazen2001segmentation,kenneth2003localization,liu2005fingerprint,yoon2013LFIQ,candela1995pcasys,jain2000filterbank,rerkrai2000new,areekul2006new,deerada2020progressive}.
However, such approaches demonstrate substantial deficiencies when confronted with incomplete or highly noisy data.
Yang \etal \cite{yang2014localized} introduced a pose estimation algorithm utilizing Hough voting.
During the offline phase, orientation fields are extracted from high-quality image patches to build region-specific dictionaries, which are then matched with input fingerprints during the online phase, with voting in Hough space and selecting the maximum response as the result.
Similarly, Yin \etal \cite{yin2018orientation} constructed a dictionary of global orientation fields from aligned high-quality fingerprints and made decision through exhaustive search.
Besides, Su \etal \cite{su2016fingerprint} employed Support Vector Machine (SVM) to build a set of classifiers for identifying fingerprint center and direction using orientation field histograms.
Furthermore, Gu \etal \cite{gu2018efficient} utilized the orientation field and periodic map of ridge patches as features, and subsequently predicted the center position and direction based on the Hough Forest model and SVM, respectively.
Despite tangible improvements, these region-based conventional machine learning approaches still underutilize available data and achieve strong performance primarily on high-quality rolled fingerprints.

Over the past decade, deep learning based data-driven approaches have achieved impressive results across diverse domains.
Ouyang \etal \cite{ouyang2017fingerprint} decomposed the pose estimation task as object detection in position and classification in rotation, and introduced Faster-RCNN as the network structure.
Schuch \etal \cite{schuch2018unsupervised} proposed an unsupervised learning paradigm, where a Siamese CNN is trained to predict the relative rotation between sample pairs and provide absolute angles during deployment.
Yin \etal \cite{yin2021joint} proposed a multi-task network which simultaneously regresses the values of center, direction, and singular points of fingerprints.
Furthermore, Arora \etal \cite{arora2023CPNet} suggested using two-stage prediction of core points through macro localization and micro-scale regression networks.
Duan \etal \cite{duan2023estimating} reformulated fingerprint pose estimation as dense prediction of grid offset vectors and employed a voting strategy.
Moreover, some researchers developed several fingerprint descriptor extraction algorithms that include spatial transformation networks (STN) \cite{engelsma2021learning,grosz2022C2CL,grosz2024AFRNet,pan2024fixed}, where the byproducts of affine transformation parameters can be considered as a form of pose representation.
These deep learning approaches have demonstrated remarkable success in processing both plane and rolled fingerprint images. 
Nevertheless, in the scenario of partial fingerprints, the substantial loss of available information presents a critical challenge that necessitates innovative approaches for effective resolution.
Fig. \ref{fig:intro} present illustrative examples that offer qualitative validation of this perspective.

On the other hand, some studies proposed estimating relative pose from paired input fingerprints \cite{he2022PFVNet,grosz2023latent,guan2025joint}.
While the accuracy is notably improved, the trade-off involves additional relative alignment steps that must be executed separately for each comparison, significantly increasing the time cost of matching. 
Furthermore, these techniques necessitate the prior enrollment of sufficient fingerprint samples to facilitate the pose estimation of new impression through comparative analysis.
Therefore, this paper focuses exclusively on absolute pose estimation methods which are more efficient and less dependent, leaving the discussion and research of such schemes for future studies.

\subsection{Pose Estimation Based on Capacitive Image}
There are many studies on predicting the three-dimensional angle (yaw and pitch, without roll) of fingers during touch from capacitance images.
Zaliva \etal \cite{zaliva2012finger} introduced multiple descriptive characteristics, such as area, centroid, and average intensity, to calculate finger angles.
Xiao \etal \cite{xiao2015estimating} further defined 42 features and used Gaussian models to regress pitch and yaw angles.
Subsequent studies \cite{mayer2017estimating,he2024TrackPose} used neural networks to directly predict finger angles from capacitive images, achieving the current SOTA performance.
Due to the lack of discriminative texture information, these approaches are primarily applicable for small angle inputs (less than $\pm90^\circ$).
In addition, inferring two-dimensional positions directly from low resolution contours (see Fig. \ref{fig:intro}) remains a huge challenge.

\section{Method}
\begin{figure*}[!t]
	\centering
	\includegraphics[width=.93\linewidth]{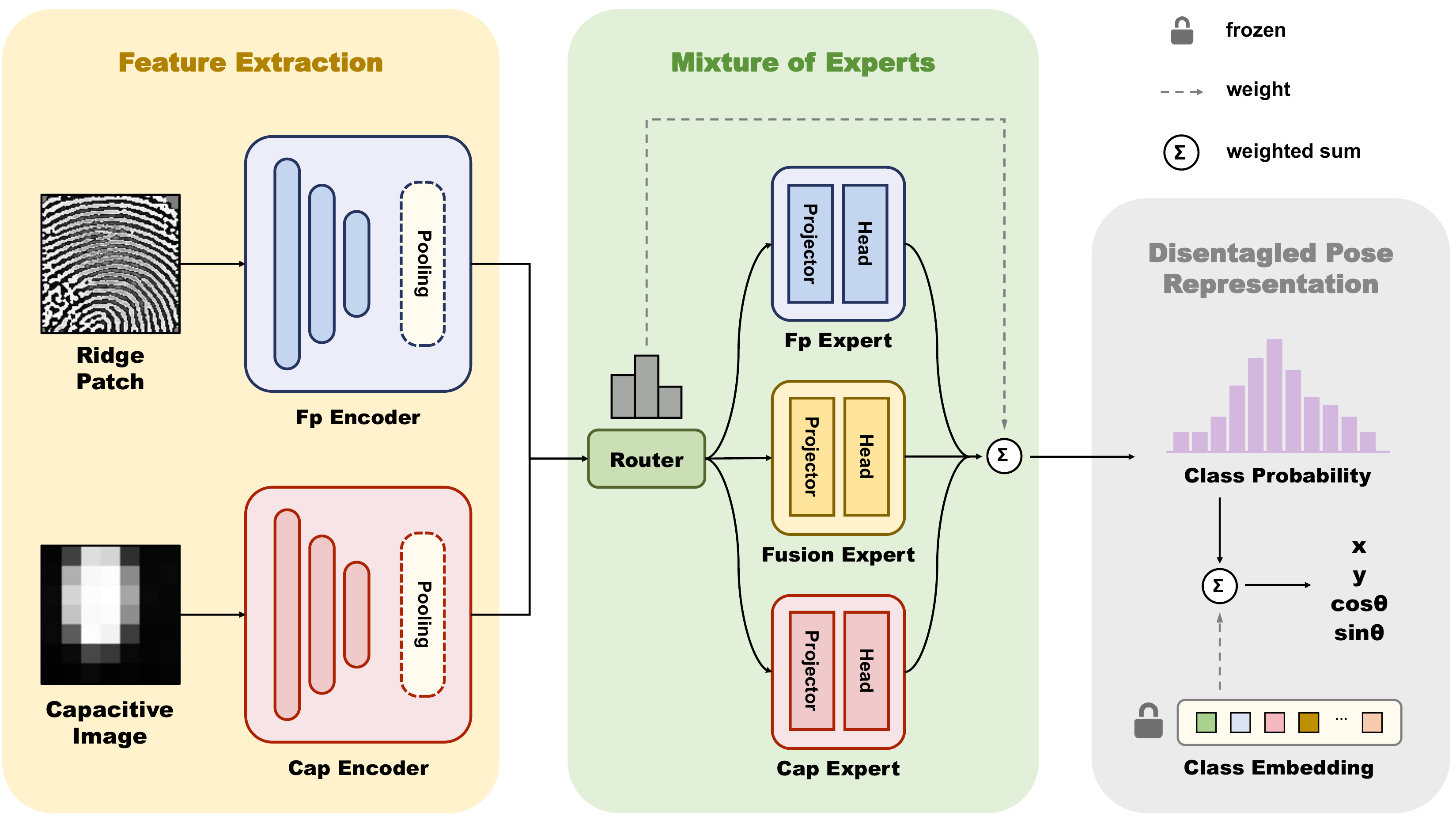}
	\caption{An overview of our partial fingerprint pose estimation network DRACO. 
    The ridge patch and capacitive image collected simultaneously from the touch device equipped with an under screen fingerprint sensor are input.
    The prediction results are represented by the horizontal and vertical coordinates of the center, as well as the sine and cosine values of the direction.
    }
	\label{fig:network}
\end{figure*}

In this section, we will specifically introduce our partial fingerprint pose estimation method, which utilizes a dual modal input consisting of ridge patches and capacitive images.
The proposed network, named DRACO, is shown in Fig. \ref{fig:network}.
Our framework employs a three-stage pipeline: 
(1) initially, two parallel branches are utilized to extract distinctive features from each modality; 
(2) subsequently, these features undergo integration through the MoE mechanism for comprehensive fusion and collaborative guidance; 
(3) ultimately, DRACO generates decoupled pose representations in the form of independent one-dimensional probability distributions.
Moreover, we design a knowledge transfer strategy from full fingerprints to partial fingerprints, as illustrated in Fig. \ref{fig:KD}, to help the network better comprehend and represent high-level semantic information as well as subtle differences between samples.
These components will be sequentially introduced, and the corresponding loss functions are presented at the end.
The construction of training data will be detailed in Section \ref{sec:dataset}.

\subsection{Feature Extraction}
In this phase, two parallel encoders with homologous structures are utilized to separately derive modality-specific features from each input stream.
Specifically, each branch begins with a stem that sequentially applies two consecutive groups of convolution, normalization, and activation operations.
Given the success of ResNeXt-34 \cite{xie2017aggregated}, we employed the same building blocks to construct a four-layer architecture with the configuration of [3, 4, 6, 3], while incorporating spatial and channel attention modules \cite{woo2018CBAM} between layers to enhance the feature representation capabilities.
It should be noted that when processing high-resolution ridge patches, the stride of each layer is set to 2 to facilitate downsampling, whereas it remains at 1 for low-resolution capacitive images.
The extracted features are ultimately converted into corresponding one-dimensional vectors via global average pooling.

\subsection{Mixture of Experts}
Inspired by \cite{jacobs1991adaptive,william2022switch,liu2024deepseek}, we introduced the MoE mechanism, which demonstrate significant efficacy in addressing the inherent heterogeneity across different domains.
The proposed architecture integrates three specialized experts: two dedicated to processing independent feature representations and one focused on mixed feature interactions, thereby flexibly enhancing the ability of multimodal feature fusion through collaborative guidance.
A straightforward router, composed of two fully connected layers, dynamically generates adaptive weights for the inference of each expert.
Let $f^\mathrm{P}$, $f^\mathrm{C}$ represent the feature vectors of corresponding branches, the processing flow can be represented as:
\begin{equation}
\begin{aligned}
    f^\mathrm{F} &=  \operatorname{cat}(\:f^\mathrm{P}, f^\mathrm{C}\:) \;,\\
   w^{\mathrm{P}},w^{\mathrm{F}},w^{\mathrm{C}} &= \operatorname{Router}[\:f^\mathrm{F} \:] \;, 
\end{aligned}
\end{equation}
where $\operatorname{cat}(\:)$ corresponds to concatenation in the channel dimension.
The subsequent feature enhancement and result assembly can be expressed as:
\begin{equation}
\begin{aligned}
   d^i  &= \operatorname{Head}^i [\:\operatorname{Proj}^i[\:f^i\:]\:] \;, \\
   d &= \sum_{{\scriptscriptstyle\mathrm{P},\mathrm{F},\mathrm{C}}} w^i \cdot d^i \;,
\end{aligned}
\end{equation}
where $d$ is corresponding pose parameter.
To effectively capture the high-level semantic features while alleviating gradient-related issues, we stack one linear layer and four Multilayer Perceptron (MLP) with residual connections \cite{touvron2023ResMLP} as projector.
On the other hand, a single linear layer is serviced as corresponding task head.

\subsection{Disentagled Pose Representation} \label{subsec:disentagled-pose-representation}
Departing from conventional numerical regression \cite{ouyang2017fingerprint,yin2021joint,engelsma2021learning,grosz2023latent,grosz2024AFRNet} or heatmap voting \cite{gu2018efficient,duan2023estimating} approaches, we reformulate pose parameters and supervision as decoupled probability distributions, thereby providing a more robust and interpretable representation.
In other words, each sample yields four pose representations as output, corresponding to the horizontal and vertical coordinates of the fingerprint center, as well as the sine and cosine of the angle.
Four sets of frozen category embeddings is pre-set to provide all spatial information.
With this assistance, the model only needs to describe the similarity between pose information and each category, rather than directly predicting the specific encoding. 
This evolution can significantly alleviate the learning complexity and enhances the generalization capability \cite{li2022SimCC,jiang2023Rtmose}.

In this paper, we divide the horizontal and vertical displacements (from $-256 \:\mathrm{px}$ to $256 \:\mathrm{px}$) into 256 segments at equal intervals, and the sine and cosine (from $-1$ to $1$) of the angle into 120 uniform segments as frozen class embeddings.
It is worth noting that trigonometric functions are used to represent fingerprint dircetion, instead of angle, to avoid confusion that may occur when approaching the two synonymous ends of $0^\circ$ and $360^\circ$.
Motivated by \cite{li2021TokenPose,li2021human}, we consider the class probability as a quantified distribution and perform weighted summation as the estimation result, rather than one-hot hard classification.
For example, given the estimated probability distribution $\{d_t\}$ and class embedding $\{e_t\}$, the horizontal coordinate $x$ of the fingerprint center can be calculated as
\begin{equation}
    x = \frac{1}{\sum_{t=1}^n d_t} \; \sum_{t=1}^n d_t \cdot e_t  \;,
    \label{eq:distribution2value}
\end{equation}
where $t$ and $n$ correspond to the index and total number of segments, respectively.
Taking this expectation helps avoid quantization errors caused by segment partitioning.
The other pose parameters $y$,$\cos \theta$ and $\sin \theta$ are also obtained according to Equation \ref{eq:distribution2value}.
Additionally, the final direction $\theta$ is obtained by calculating the arctangent of these two trigonometric functions.

\subsection{Knowledge Transfer} \label{subsec:knowledge-transfer}

\begin{figure}[!t]
    \centering
    \includegraphics[width=0.95\linewidth]{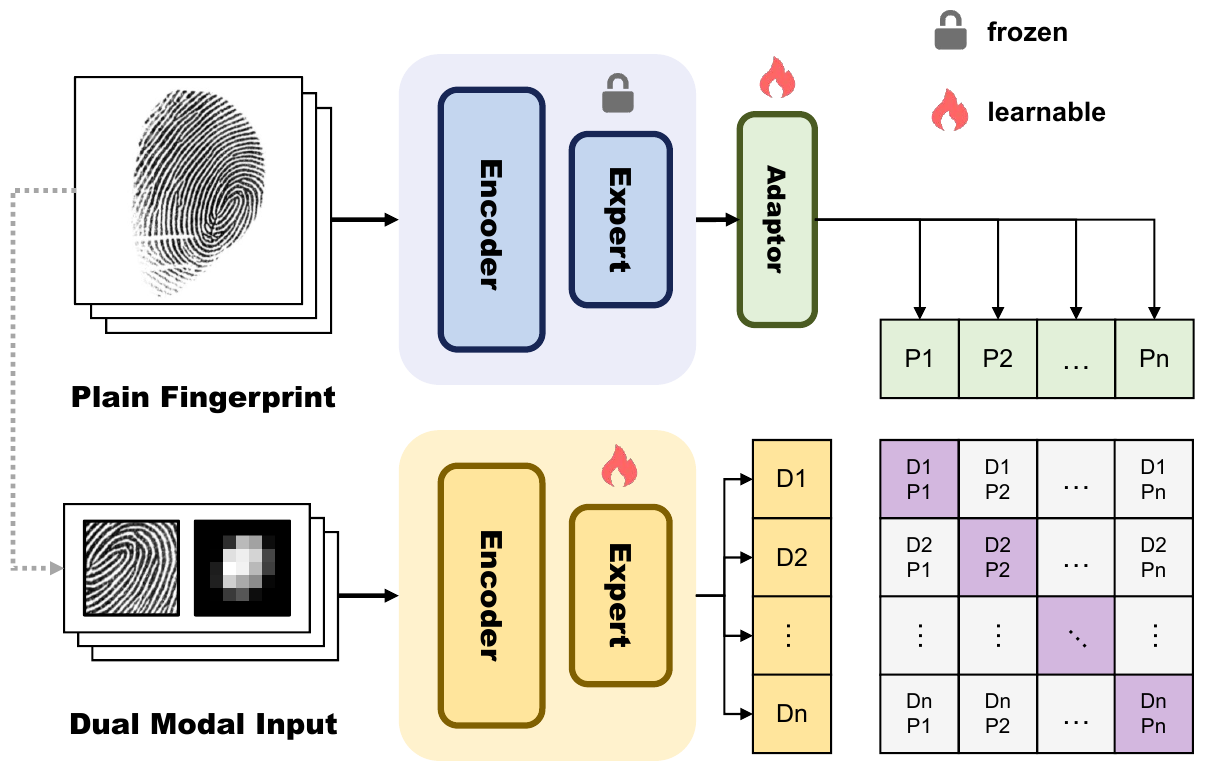}
    \caption{The main process of knowledge transfer during network training. 
    The blue and yellow modules corresponds to modules of the same color in Fig. \ref{fig:network}.
    As shown in the bottom right corner, the features of dual modal inputs are progressively aligned with the depiction in plain fingerprints through contrastive learning techniques.
    The gray dashed line represents the process of data synthesis.
    }
    \label{fig:KD}
\end{figure}

During the experiments, we observed that the same model demonstrated significantly superior performance on full fingerprint compared to partial fingerprint modalities.
This phenomenon is reasonable because full fingerprints contain almost all the available one-sided features from both ridge patches and capacitive images, as well as richer and more comprehensive texture information.
Therefore, we employ contrastive learning techniques \cite{grill2020BYOL,radford2021CLIP} to infuse global prior correlations (in plain fingerprint) into the current model applied to degraded modalities (partial fingerprint), leveraging a relationship-based knowledge transfer strategy \cite{zhou2023UniDistrill,yang2024CLIPKD}.
As shown in Fig. \ref{fig:KD}, a teacher model (blue) is pre-trained on plain fingerprints and its parameters are frozen.
Next, a dual-modal student (yellow) is trained with the objective of extracting features whose relative relationships align as closely as possible with the judgments of the teacher.
It should be noted that there is a strong correspondence between the dual modal input and plain fingerprint in each group, as the former is simulated from the latter and its process details are introduced in Section \ref{sec:dataset}.
In this paper, we use a network with the same structure as the fingerprint part in Fig. \ref{fig:network} as teacher, and reinforce the corresponding feature extraction and fusion expert through this task.
Specifically, a stacked 2-layer MLP is used as adapter to appropriately adjust the inherent information gap between teacher and student.
Under this relationship based supervision, features from different modals of the same impression are brought as close as possible, while features from distinct impressions are intentionally separated to maximize their distance.
By leveraging these relational insights, the model gains a deeper understanding of data nuances, resulting in better outcomes without any additional cost during the testing stage.

\subsection{Loss Function}
For convenience, we jointly optimize the pose estimation and knowledge transfer process in one training process.
As outlined in Section \ref{subsec:disentagled-pose-representation}, we employ the distance between quantitative probability distributions associated with each parameter as the supervisory signal for pose estimation task.
Let $\mathrm{\Phi}$ represent a certain pose component, $d$ and $\tilde{d}$ represent the predicted result and ground truth of corresponding probability distribution, the loss function of pose estimation $\mathcal{H}$ is defined as:
\begin{equation}
    \mathcal{H} = \sum_{\mathrm{\Phi} \in \{\mathrm{x},\mathrm{y},\mathrm{cos},\mathrm{sin} \}} \lambda_\mathrm{\Phi} \cdot \operatorname{dist} (\: d_\mathrm{\Phi},  \tilde{d}_\mathrm{\Phi} \:) \;, 
\end{equation}
where cross entropy (CE) serves as the distance metric $\operatorname{dist}(\:)$ between two distributions.
All balance factors $\lambda$ are empirically set to $1.0$.
The value $\tilde{v}$ of ground truth is converted into discrete probability $\tilde{d}$ using gaussian distribution:
\begin{equation}
    \tilde{d}_t=\exp \left(-\frac{\tilde{v}-e_t}{2 \sigma^2}\right) / \sum_t \tilde{d}_t \;,
\end{equation}
where the definition of $e_t$ is consistent with Equation \ref{eq:distribution2value}, and the hyperparameter $\sigma$ is set to 3.5 and 2.5 for position and angle sub-losses.
To further augment the individual capabilities of each expert, we integrate classification heads with the same structure after each expert branch as subtasks.
The total loss of the pose estimation part is
\begin{equation}
    \mathcal{L}_{\mathrm{pose}} = \sum \lambda_{e} \cdot \mathcal{H}_{e} \;, 
    \label{eq:loss-pose}
\end{equation}
where the hyperparameters $\lambda_{e}$ corresponding to the three experts (Fp, Fusion and Cap in Fig. \ref{fig:network}) and the comprehensive results, fixed as 0.2, 0.2, 0.4, and 1.0 respectively.

On the other hand, the Information Noise Contrastive Estimation (InfoNCE) with temperature coefficients is used to optimize the knowledge distillation process in Section \ref{subsec:knowledge-transfer}.
Referring to Fig. \ref{fig:KD}, a plain fingerprint and the corresponding simulated dual modal images are used as the input group during training.
For the features $\{P\}$ and $\{D\}$ extracted by the teacher and student networks, this relationship-based supervision is  computed as:
\begin{equation}
    \begin{aligned}
        z(D_i,P) &=\frac{\exp \left(\operatorname{sim}\left(D_i, D_i^{+}\right) / \tau\right)}{\sum_{j=1}^B \exp \left(\operatorname{sim}\left(D_i, P_j\right) / \tau\right)} \;,\\
        \mathcal{L}_{\mathrm{KT}} &= - \frac{1}{2B} \sum_{i} \left(\:\log z(D_i,P) + \log z(P_i,D) \:\right) \;,
    \end{aligned}
\end{equation}
where $B$ is the batch size, $\operatorname{sim}$ is the cosine similarity.
The temperature $\tau$ is set to 8.0 according to the results of small-scale parameter search.

The final loss is computed as the weighted sum of aforementioned two tasks:
\begin{equation}
    \mathcal{L} = \mathcal{L}_{\mathrm{pose}} + \lambda \cdot \mathcal{L}_{\mathrm{KT}} \;.
\end{equation}
The balance term $\lambda$ is configured to 1.0.

\section{Dataset} \label{sec:dataset}

\subsection{Dataset Introduction}

\begin{table*}[!t]
	\renewcommand\arraystretch{1.3}
	\caption{All Fingerprint Datasets Used in Experiments. 
    Among them, partial fingerprint refers to two modalities: ridge patches and capacitive images.}
	\label{tab:dataset}
	\centering
    \begin{threeparttable}
        \setlength{\tabcolsep}{4pt}
        \begin{tabular*}{1.0\textwidth}{@{\extracolsep{\fill}} c c c c c c @{}}
            \toprule
          \textbf{Dataset} 
            & \textbf{Type} 
            & \textbf{Description}              
            & \textbf{Usage}        
            & \textbf{Genuine pairs\tnote{\,a}}
            & \textbf{Impostor pairs\tnote{\,a}} \\
            \midrule
            FVC2002 DB1\_A \cite{fvc2002}
            & Plain fingerprints with front pose
            & 100 fingers $\times$ 8 impressions
            & test
            & n/a
            & n/a \\
            FVC2004 DB1\_A \cite{fvc2004}
            & Plain fingerprints with front pose
            & 100 fingers $\times$ 8 impressions
            & test
            & n/a
            & n/a \\
            DPF \cite{duan2023estimating}
            & Rolled fingerprints
            & 933 fingers $\times$ 1 impression
            & calibration
            & n/a
            & n/a \\
            {}
            & Plain fingerprints with diverse poses\tnote{\,b}
            & 933 fingers $\times$ 3.1 impressions\tnote{\,c}
            & train \& test
            & 40,579
            & 4,157,822 \\
            PCF 
            & Rolled fingerprints
            & 100 fingers $\times$ 1 impression
            & calibration
            & n/a
            & n/a \\
            {}
            & Partial fingerprints with diverse poses
            & 100 fingers $\times$ 32 impression
            & finetune \& test
            & 46,338
            & 3,413,262 \\
            \bottomrule
        \end{tabular*}
        \begin{tablenotes}
            \item[a] Effective pairs in matching experiments.
            \item[b] Simultaneously, partial fingerprints are simulated based on plain fingerprints.
            \item[c] Average number after screening.
        \end{tablenotes}
    \end{threeparttable}
    
\end{table*}

\begin{figure*}[!t]
	\centering
	\subfloat[]{\includegraphics[height=.13\linewidth]{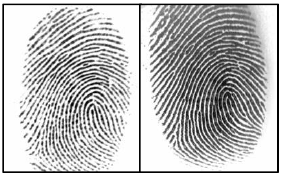}%
	\vspace{-1mm}}
	\hfil
	\subfloat[]{\includegraphics[height=.13\linewidth]{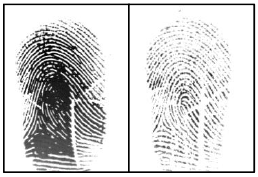}%
	\vspace{-1mm}}
	\hfil
	\subfloat[]{\includegraphics[height=.13\linewidth]{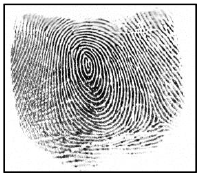}%
	\vspace{-1mm}}
	\hfil
	\subfloat[]{\includegraphics[height=.13\linewidth]{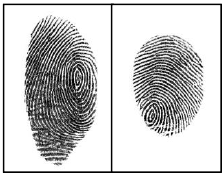}%
	\vspace{-1mm}}
	
	\par
	
	\subfloat[]{\includegraphics[height=.13\linewidth]{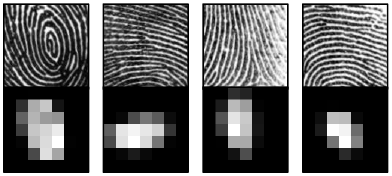}%
	\vspace{-1mm}}
	\hfil
	\subfloat[]{\includegraphics[height=.13\linewidth]{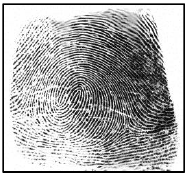}%
	\vspace{-1mm}}
	\hfil
	\subfloat[]{\includegraphics[height=.13\linewidth]{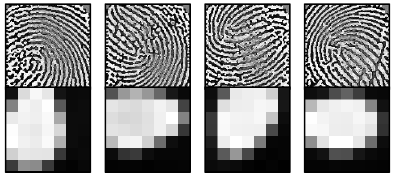}%
	\vspace{-1mm}}
	\caption{Image examples from different datasets: (a) FVC2002 DB1\_A \cite{fvc2002}, (b) FVC2004 DB1\_A \cite{fvc2004}, (c) DPF \cite{duan2023estimating}, rolled fingerprint, (d) DPF \cite{duan2023estimating}, plain fingerprints, (e) DPF \cite{duan2023estimating}, simulated partial fingerprints, (f) PCF, rolled fingerprints, (g) PCF, partial fingerprints.
    The `partial fingerprint' is a general term used to represent two modalities: ridge patch and capacitive image.
    }
	\label{fig:dataset}
\end{figure*}

The utilization information of all datasets are listed in Table \ref{tab:dataset}.
Moreover, Fig. \ref{fig:dataset} shows corresponding image examples.
Considering that most existing public datasets are predominantly collected in frontal poses and lack diversity in terms of location, we only used two common datasets (\emph{FVC2002 DB1\_A} \cite{fvc2002} and \emph{FVC2004 DB1\_A} \cite{fvc2002}) as representatives.
Private \emph{DPF} database \cite{duan2023estimating} explicitly require subjects to adopt diverse poses during collection, making it more suitable for this task.
Specifically, 8,479 samples from 744 fingers were used for training, while the remaining 2,049 samples (from other 189 fingers) were allocated for testing.
In addition, we collected and established a real partial fingerprint dataset using a smartphone equipped with an underscreen fingerprint sensor.
In this paper, it referred to as the Phone Captured Fingerprint and Rolled Fingerprint Database (\emph{PCF}).
Ridge patches with a size of $132\times132\:\mathrm{px}$ in 500 ppi, as well as capacitive images with $7\times7\:\mathrm{px}$ (effective area) in 10 ppi, can be obtained simultaneously.
A total of 640 images from 20 fingers were used to fine-tune the model, and 2,560 images from other 80 fingers were used for testing.
In addition, subjects in \emph{DPF} and \emph{PCF} were also required to collect rolled fingerprints, which were used for subsequent calibration to obtain pose ground truth.

\subsection{Training Set Construction} \label{subsec:training-set-construction}
We use the subset of \emph{DPF} mentioned above for training.
For full fingerprint scenarios, we directly utilize the samples displayed in Fig. \ref{fig:dataset}(d) for training. 
The ground truth is obtained by minutiae matching between the input plain fingerprint and its corresponding pose-standardized rolled fingerprint.
In this paper, we employ VeriFinger SDK 12.0 \cite{verifinger} for minutiae extraction and matching, and adopt the method proposed by Duan \etal \cite{duan2023estimating} for rolled fingerprint pose rectification.
These approaches are proven to be sufficiently reliable under conditions of high image quality and small pose variation.
Therefore, we approximately consider it as an unbiased and precise calibration.

Due to the scarcity of readily available large-scale partial fingerprint data and capacitive images, we generated approximate samples from the plain fingerprints of \emph{DPF}.
Following \cite{he2022PFVNet,duan2023finger,guan2025joint}, we randomly cropped square regions from the plain fingerprint to simulate ridge patches. 
On the other hand, window-based uniform filtering and interpolation are applied to downsample the original image to 10 ppi, emulating the capacitive image \cite{duan2023finger}.
The setting of cropping size and target resolution is to maintain consistency with the real data in \emph{PCF}.
Nevertheless, we fine-tuned our model using a small amount of local data before testing on \emph{PCF} to minimize domain bias as much as possible.

\subsection{Test Set Protocol}
To comprehensively evaluate the accuracy and robustness of pose estimation algorithms, we generated four distinct test sets for each scenarios.
Specifically, the pose of samples in \emph{FVC2002 \& FVC2004 DB1\_A} and \emph{DPF} are first standardized and then randomly rotated within the four ranges of $\pm 45^\circ$, $\pm 90^\circ$, $\pm 135^\circ$ and $\pm 180^\circ$.
Each evaluated method will be trained and tested separately based on the range of direction angle under full fingerprint or partial fingerprint scenario.
For simplicity, the training conditions of each model will not be specifically declared in experiments.
Additionally, the real partial fingerprint data \emph{PCF} remains unchanged in order to accurately provide feedback on the real environment.
The ground truth of its pose information is obtained through the same calibration process introduced in Section \ref{subsec:training-set-construction}.
In the matching experiments, pairs with the same identity will be checked in advance to exclude situations where there is no overlapping area.
The effective number of genuine and impostor pairs is presented in Table \ref{tab:dataset}.

\section{Experiments}
In experiments, we compare the proposed DRACO with SOTA finger pose estimation algorithms on full fingerprints (plain fingerprints) and partial fingerprints (ridge patches and capacitive images).
The implementation details of DRACO are provided first, followed by an introduction to the representative methods used for comparison.
The performance of these algorithms is then thoroughly evaluated in terms of pose estimation and matching capabilities across various rotation ranges. 
This assessment ensures a comprehensive understanding of how each algorithm performs under different conditions, highlighting their strengths and weaknesses in handling pose variations.
In addition, extensive ablation experiments are conducted to validate the effectiveness of our proposed modules and strategies, while also offering potential inspiration for future works.
Finally, the efficiency of different algorithms is reported to assess their deployment costs.

\subsection{Implementation Details}
Our proposed DRACO are trained under the corresponding modality in \emph{DPF} with an initial learning
rate of  $1\mathrm{e}{-3}$ (end of $1\mathrm{e}{-6}$), cosine annealing scheduler, default AdamW optimizer and batch size of 256 for 80 epochs. 
Data augmentation is used, including random translation within $\pm 40$ pixels and random rotation of $\pm45^\circ$, $\pm90^\circ$, $\pm135^\circ$, and $\pm180^\circ$ (according to the corresponding test scenario).
Specifically, when incorporating contrastive learning, we increase the batch size to 512 to ensure sample richness.
The learning rate and epoch number is adjusted to $4\times$ and 200 to roughly maintain the original optimization iterations.
Before testing on \emph{PCF}, the parameters under $\pm180^\circ$ augmentation are loaded and further fine-tuned with an initial learning rate of  $1\mathrm{e}{-4}$ (end of $1\mathrm{e}{-5}$) for 200 epochs, and keep other parameters consistent.
When DRACO is applied in a single modal, only the corresponding feature extractor and expert shown in Fig. \ref{fig:network} are activated.
In following experiments, we used suffixes `fp' and `cap' to distinguish models with DRACO structures that only use branch of ridge patch and capacitive image, respectively.

\subsection{Compared Methods}
For plain fingerpint and ridge patch, we reproduced four representative methods, including: 
\begin{itemize}
\item \textbf{Faster-RCNN}: Object detection network proposed by Ouyang \etal \cite{ouyang2017fingerprint};
\item \textbf{STN}: STN module under indirect supervision of fixed-length representation task \cite{engelsma2021learning,grosz2024AFRNet};
\item \textbf{JointNet}: Numerical regression network proposed by Yin \etal \cite{yin2021joint};
\item \textbf{GridNet}: Heatmap voting network recently proposed by Duan \etal \cite{duan2023estimating}.
\end{itemize}
Additionally, for the capacitive image modality, since existing researches focus on predicting angles in 3D space, we re-implement the following approaches and adjust the output head to estimate the center and rotation value of 2D pose:
\begin{itemize}
\item \textbf{Cap-MLP}: Regressor based on manually defined features and multi-layer MLP, inspired by Xiao \etal \cite{ouyang2017fingerprint};
\item \textbf{Cap-CNN}:Numerical regression network inspired by recent works \cite{mayer2017estimating,he2024TrackPose}.
\end{itemize}
Above methods are also trained under the combinations of settings in DRACO to ensure sufficient fairness in the comparison.

\subsection{Pose Estimation Performance} \label{subsec:pose-estimation-performance}
Although the focus of this paper is on partial fingerprint pose estimation, our disentagled pose representation strategy is universal. 
Therefore, we first compare representative methods with our proposed DRACO in full fingerprint scenario, and then provide detailed evaluation on partial fingerprint (including partial fingerprint and capacitive image) pose estimation.

\subsubsection{Evaluation on Full Fingerprints}
\begin{figure*}[!t]
	\centering
	\subfloat[]{\includegraphics[width=.24\linewidth]{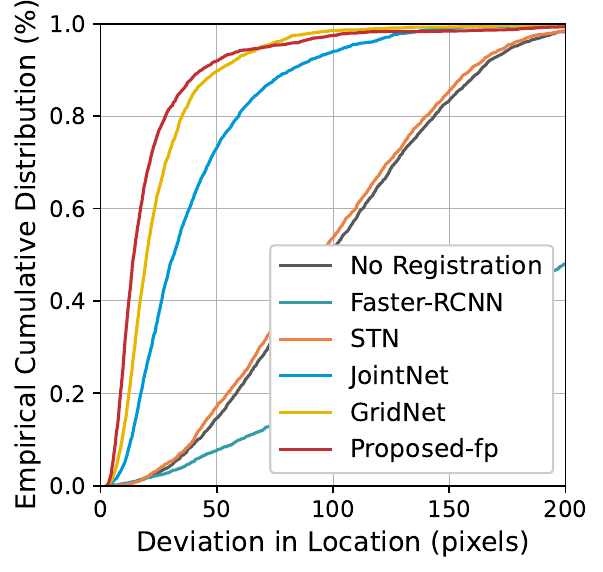}%
	\vspace{-1.5mm}}
	\hfil
	\subfloat[]{\includegraphics[width=.24\linewidth]{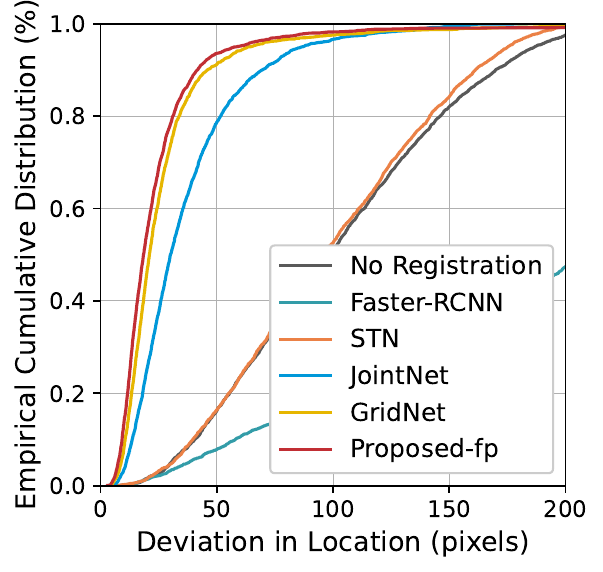}%
	\vspace{-1.5mm}}
    \hfil
    \subfloat[]{\includegraphics[width=.24\linewidth]{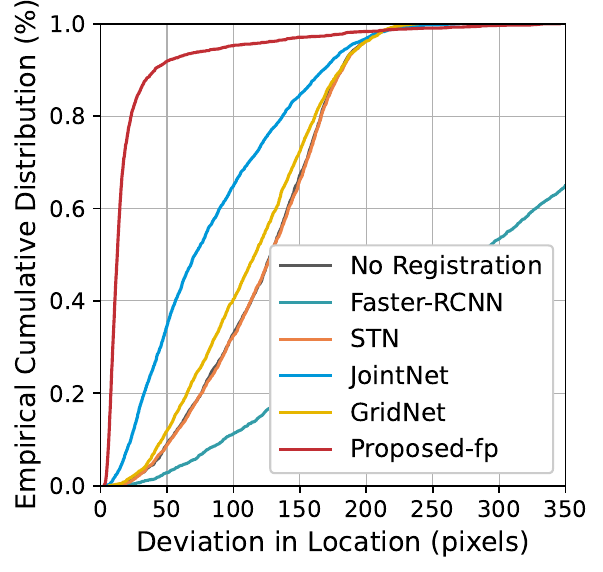}%
	\vspace{-1.5mm}}
	\hfil
	\subfloat[]{\includegraphics[width=.24\linewidth]{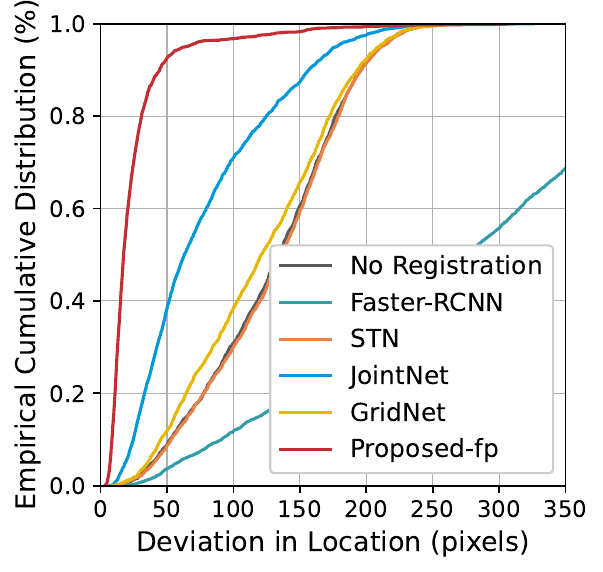}%
	\vspace{-1.5mm}}

	\caption{The empirical cumulative distribution of location deviations on  \fbox{full fingerprints} from (a) FVC2002 DB1\_A $[-90^\circ,90^\circ]$, (b) FVC2004 DB1\_A $[-90^\circ,90^\circ]$, (c) FVC2002 DB1\_A $[-180^\circ,180^\circ]$, (d) FVC2004 DB1\_A $[-180^\circ,180^\circ]$.
    Suffixes `fp' indicate that only the corresponding branch is used.}
	\label{fig:ECD-trans}
\end{figure*}

\begin{figure*}[!t]
	\centering
	\subfloat[]{\includegraphics[width=.24\linewidth]{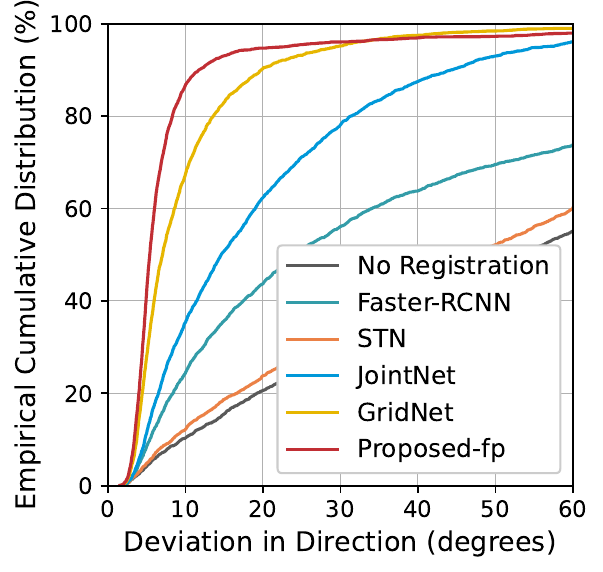}%
	\vspace{-1.5mm}}
	\hfil
	\subfloat[]{\includegraphics[width=.24\linewidth]{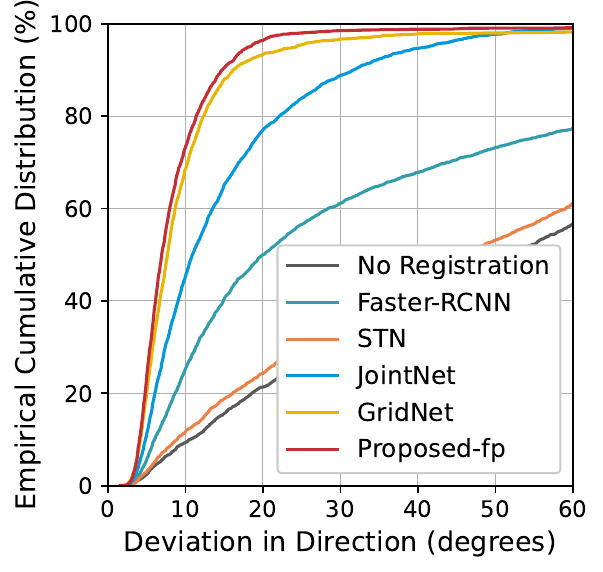}%
	\vspace{-1.5mm}}
    \hfil
    \subfloat[]{\includegraphics[width=.24\linewidth]{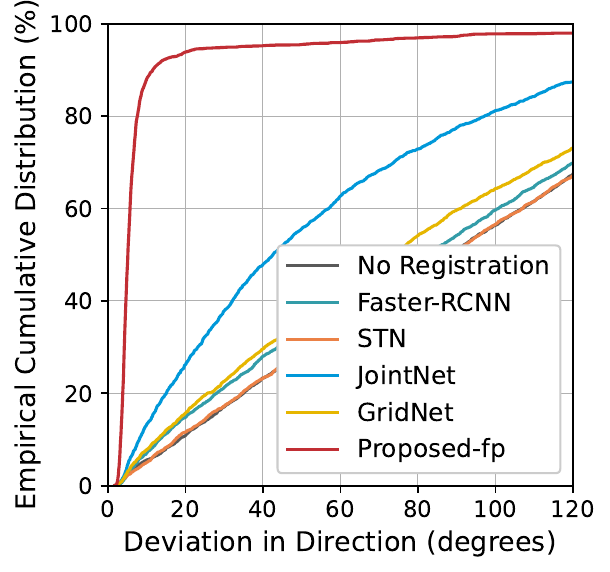}%
	\vspace{-1.5mm}}
	\hfil
	\subfloat[]{\includegraphics[width=.24\linewidth]{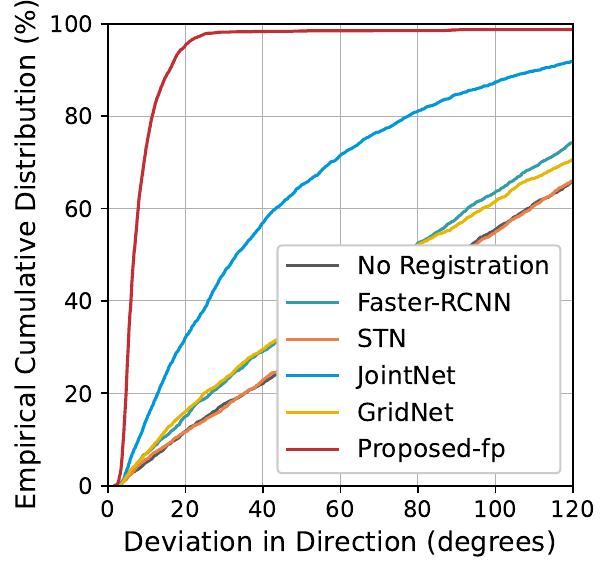}%
	\vspace{-1.5mm}}
	
	\caption{The empirical cumulative distribution of direction deviations on \fbox{full fingerprints} from  (a) FVC2002 DB1\_A $[-90^\circ,90^\circ]$, (b) FVC2004 DB1\_A $[-90^\circ,90^\circ]$, (c) FVC2002 DB1\_A $[-180^\circ,180^\circ]$, (d) FVC2004 DB1\_A $[-180^\circ,180^\circ]$.
    Suffixes `fp' indicate that only the corresponding branch is used.}
	\label{fig:ECD-rot}
\end{figure*}

Consistent with previous works \cite{yin2021joint,duan2023estimating}, the deviation in location and direction of mated minutiae pairs on \emph{FVC2002 DB1\_A} and \emph{FVC2004 DB1\_A} is presented to characterize the accuracy of pose estimation.
Specifically, each method infers the 2D pose of single plain fingerprint, which is then utilized to execute a rigid transformation and align the image to standard coordinate system.
Subsequently, VeriFinger \cite{verifinger} is used to extract and match minutiae between impressions of the same identity.
The average deviation of Euclidean distance and absolute rotation error between paired points is visualized through empirical cumulative score function.
As shown in Fig. \ref{fig:ECD-trans} and Fig. \ref{fig:ECD-rot}, the proposed DRACO exhibits significant advantages across all test datasets, especially when dealing with large rotation angles.
The performance of \emph{STN} \cite{engelsma2021learning,grosz2024AFRNet} is close to no registration, primarily because pose estimation is not directly supervised. 
This absence of explicit guidance may impede the network's ability to learn accurate transformations.
In addition, the object detection based \emph{Faster-RCNN} \cite{ouyang2017fingerprint} shows limited performance in estimating fingerprint position. 
A compelling explanation is that in this task, the network primarily focuses on the outer contours, which may lead to neglect of the precise center.

Furthermore, we evaluated the performance of these algorithms across four different rotation ranges on plain fingerprints of \emph{DPF}.
The mean absolute error of translation and rotation is reported in Table \ref{tab:pose-plain}.
It can be seen that the performance of previous SOTA methods significantly declines as the rotation range increases.
In contrast, our method demonstrates exceptional precision and robustness across the entire rotation range.
Based on these experimental results, we selected the best-performing algorithms, \emph{JointNet} \cite{yin2021joint} and \emph{GridNet} \cite{duan2023estimating}, for the subsequent experiments on partial fingerprints.

\subsubsection{Evaluation on Partial Fingerprints}

\begin{table*}[!t]
	\renewcommand\arraystretch{1.3}
	\belowrulesep=-0.2pt
	\aboverulesep=-0.2pt
	\caption{
    Alignment error under different fingerprint poses on DPF (\fbox{partial fingerprints} with different rotation ranges).
    The four groups, from top to bottom, represent default configuration and methods that use only ridge patches, only capacitive images, and a combination of both.
    Suffixes `fp' and `cap' indicate that only the corresponding branch is used.
    Bold and underlined number represent the corresponding global optimal result and optimal method in each group respectively.
    }
	\label{tab:pose-partial-dpf}
	\vspace{-0.4cm}
	\begin{center}
		\begin{threeparttable}
				\setlength{\tabcolsep}{12pt}
                \begin{tabular}{c | c c | c c | c c | c c}
                \toprule
                \multirow{2}{*}[-.0mm]{\textbf{Method}} 
                & \multicolumn{2}{c|}{{\textbf{[$\mathbf{-45^\circ,45^\circ}$]}}}
                & \multicolumn{2}{c|}{{\textbf{[$\mathbf{-90^\circ,90^\circ}$]}}}
                & \multicolumn{2}{c|}{{\textbf{[$\mathbf{-135^\circ,135^\circ}$]}}}
                & \multicolumn{2}{c}{{\textbf{[$\mathbf{-180^\circ,180^\circ}$]}}}
                \\
                \cmidrule(lr){2-3}\cmidrule(lr){4-5}\cmidrule(lr){6-7}\cmidrule(lr){8-9}
                {}
                & \scriptsize\textbf{trans (px)} & \scriptsize\textbf{rot ($^\circ$)}
                & \scriptsize\textbf{trans (px)} & \scriptsize\textbf{rot ($^\circ$)}
                & \scriptsize\textbf{trans (px)} & \scriptsize\textbf{rot ($^\circ$)}
                & \scriptsize\textbf{trans (px)} & \scriptsize\textbf{rot ($^\circ$)}
                \\
                \midrule
                No Registration
                & 83.4 & 22.7
                & 85.5 & 45.5
                & 85.1 & 67.4
                & 88.3 & 90.8
                \\
                \hline
                JointNet \cite{yin2021joint}
                & 34.3 & 17.3
                & 36.1 & 18.7
                & 37.3 & 27.9
                & 41.4 & 34.5
                \\
                GridNet \cite{duan2023estimating}
                & 35.2 & 12.0
                & 44.2 & 19.6
                & 47.3 & 37.8
                & 74.8 & 64.5
                \\
                \cellcolor{black!10}{Proposed-fp}
                & \cellcolor{black!10}\underline{23.4} & \cellcolor{black!10}\underline{10.3}
                & \cellcolor{black!10}\underline{22.6} & \cellcolor{black!10}\underline{11.7}
                & \cellcolor{black!10}\underline{26.8} & \cellcolor{black!10}\underline{15.1}
                & \cellcolor{black!10}\underline{25.4} & \cellcolor{black!10}\underline{14.1}
                \\
                \hline
                Cap-MLP \cite{xiao2015estimating}
                & 79.7 & 13.3
                & 79.3 & 39.2
                & 83.5 & 65.5
                & 86.2 & 91.1
                \\
                Cap-CNN \cite{mayer2017estimating,he2024TrackPose}
                & 67.8 & 7.8
                & \underline{66.2} & 16.0
                & 71.7 & 58.5
                & 78.3 & 70.4
                \\
                \cellcolor{black!10}{Proposed-cap}
                & \cellcolor{black!10}\underline{65.0} & \cellcolor{black!10}\underline{7.1}
                & \cellcolor{black!10}67.8 & \cellcolor{black!10}\underline{12.9}
                & \cellcolor{black!10}\underline{70.8} & \cellcolor{black!10}\underline{49.7}
                & \cellcolor{black!10}\underline{75.7} & \cellcolor{black!10}\underline{68.1}
                \\
                \hline
                \rowcolor{black!10}
                \cellcolor{black!10}{Proposed}
                & \cellcolor{black!10}\textbf{18.4} & \cellcolor{black!10}\textbf{4.8}
                & \cellcolor{black!10}\textbf{19.5} & \cellcolor{black!10}\textbf{5.4}
                & \cellcolor{black!10}\textbf{20.3} & \cellcolor{black!10}\textbf{5.5}
                & \cellcolor{black!10}\textbf{19.8} & \cellcolor{black!10}\textbf{5.5}
                \\
                
                \bottomrule
                \end{tabular}
		\end{threeparttable}
	\end{center}
\end{table*}

\begin{table}[!t]
	\renewcommand\arraystretch{1.3}
	\belowrulesep=-0.2pt
	\aboverulesep=-0.2pt
	\caption{
    Alignment error under different fingerprint poses on  \fbox{partial fingerprints} from PCF.
    The grouping rules are the same as Table \ref{tab:pose-partial-dpf}.
    }
	\label{tab:pose-partial-pcf}
	\vspace{-0.4cm}
	\begin{center}
		\begin{threeparttable}
				\setlength{\tabcolsep}{18pt}
                \begin{tabular}{c | c c }
                \toprule
                \textbf{Method}
                & \scriptsize\textbf{trans (px)} & \scriptsize\textbf{rot ($^\circ$)}
                \\
                \midrule
                No Registration
                & 98.4 & 82.0
                \\
                \hline
                JointNet \cite{yin2021joint}
                & 51.5 & 38.0
                \\
                GridNet \cite{duan2023estimating}
                & 87.2 & 63.2
                \\
                \hline
                Cap-MLP \cite{xiao2015estimating}
                & 94.8 & 75.9
                \\
                Cap-CNN \cite{mayer2017estimating,he2024TrackPose}
                & 90.6 & 68.2
                \\
                \hline
                \rowcolor{black!10}
                \cellcolor{black!10}{Proposed}
                & \cellcolor{black!10}\textbf{31.2} & \cellcolor{black!10}\textbf{16.3}
                \\
                
                \bottomrule
                \end{tabular}
		\end{threeparttable}
	\end{center}
\end{table}

Similarly, we assessed the performance of pose estimation algorithms on partial fingerprints (partial fingerprints and capacitive images) using the simulated test set from \emph{DPF}.
As demonstrated in Table \ref{tab:pose-partial-dpf}, our proposed solution outperforms previous methods in respective unimodal groups.
This result strongly emphasizes the advantages of our pose representation scheme once again.
In addition, the full version of DRACO showcases comprehensive and notable leadership in both accuracy and stability following the integration of dual modal information.
This substantial improvement clearly demonstrates the considerable complementarity between ridge patches and capacitive images, affirming the effectiveness of the collaborative dual-modal guidance paradigm.
Tabel \ref{tab:pose-partial-pcf} presents the key comparative results on \emph{PCF}, which also highlight the impressive success of our proposed algorithm.
The overall performance on \emph{PCF} is somewhat inferior to the simulation dataset of \emph{DPF}, possibly due to domain bias and the limited fine-tuning data (640 samples during finetuning, which is significantly smaller than the 8,479 samples in training stage).
However, this experiment still provide valuable insights into evaluating the relative performance of pose estimation algorithms, which is our primary concern.

\subsubsection{Visual Analysis}
\begin{figure*}[!t]
	\centering
	\includegraphics[width=.85\linewidth]{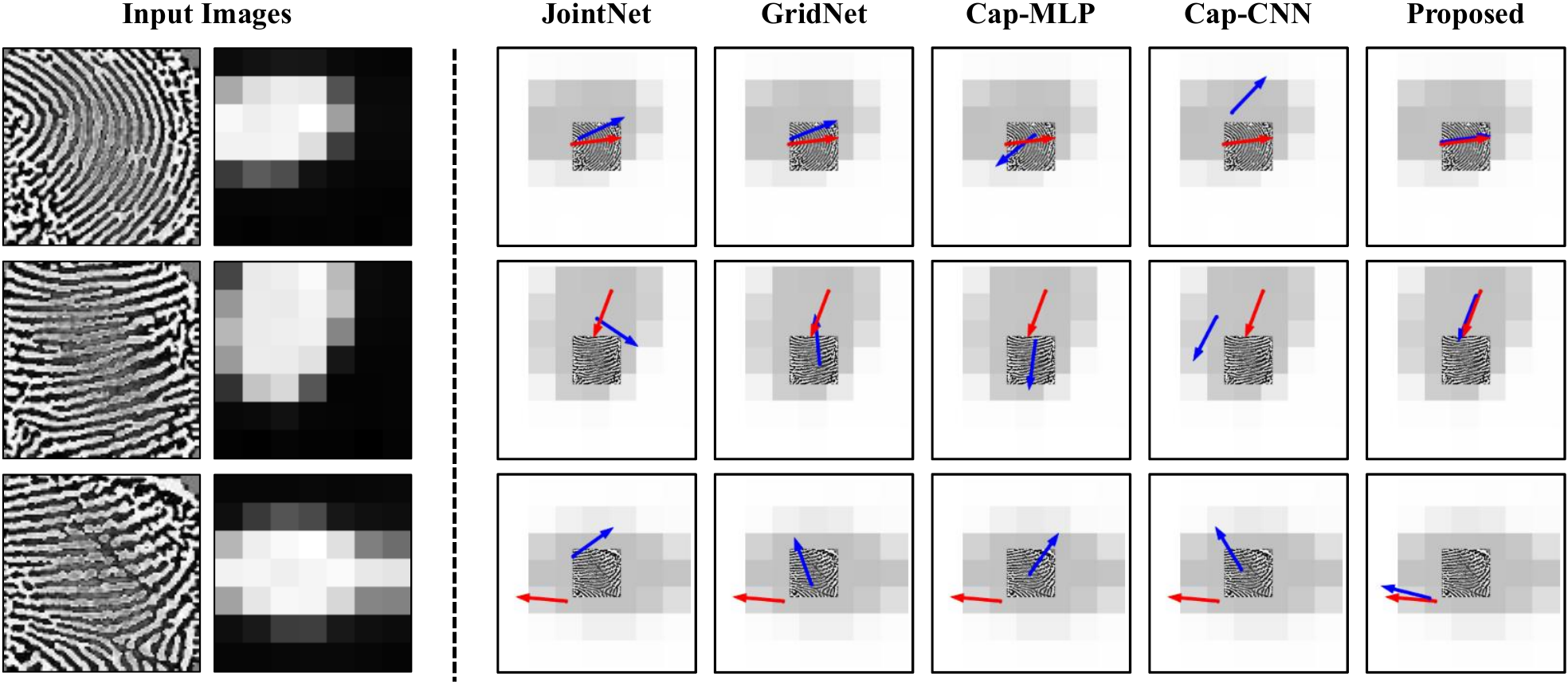}
	\caption{Examples of different pose estimation methods on  \fbox{partial fingerprints} from \emph{PCF}.
    To facilitate observation, the capacitive image is inverted and overlayed as background onto the corresponding ridge patch, displaying both the prediction result (blue arrow) and ground truth (red arrow).
    }
	\label{fig:example}
\end{figure*}

Several intuitive examples are provided to qualitatively compare different pose estimation algorithms.
Fig. \ref{fig:example} shows three representative visualization results.
It can be observed that fingerprint based methods (\emph{JointNet} \cite{yin2021joint},\emph{GridNet} \cite{duan2023estimating}) function effectively when the texture features of ridge patches possess sufficient discriminability (line 1).
On the other hand, it is possible to accurately infer angles using only capacitive images (\emph{Cap-MLP} \cite{xiao2015estimating}, \emph{Cap-CNN} \cite{mayer2017estimating,he2024TrackPose}), which  is significantly ahead of relying solely on ridge patches (line 2).
However, capacitive image based methods expose obvious deficiencies in localization.
The respective advantages and limitations of these two modals effectively highlight their complementarity.
Naturally, our method, guided by dual-modal collaboration, demonstrates more precise performance (line 1 \& 2).
Even when previous solutions have completely failed, it can still provide accurate predictions (line 3).

\begin{figure}[!t]
	\centering
	\includegraphics[width=.85\linewidth]{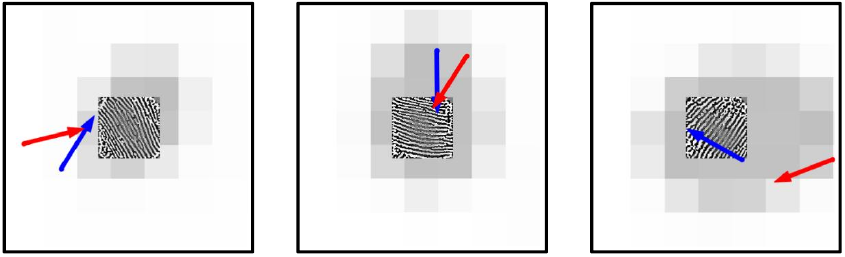}
	\caption{Failure cases of DRACO on \fbox{partial fingerprints} from \emph{PCF}. 
    The visualization protocol is consistent with Fig. \ref{fig:example}.
    }
	\label{fig:failure}
\end{figure}

We further illustrate some failure cases of DRACO in Fig. \ref{fig:failure}.
Under certain extreme pressing postures, such as those involving fingertips (column 1 \& 2), the model may encounter substantial pose estimation errors.
Additionally, in rare instances where both ridge patches and capacitive images lack sufficient recognition, DRACO may occasionally experience considerable confusion and misjudgment. (column 3).

\subsection{Matching Performance}

Fingerprint pose estimation serves as a valuable source of auxiliary information in matching tasks \cite{su2016fingerprint,maltoni2022handbook,duan2023estimating}. 
For instance, an impression taken from the left side of a finger should not be considered a successful match with any impression from the right side, no matter how similar they are.
According to this logic, recognition systems can swiftly identify and discard candidate samples that exhibit incompatible positions and orientations. 
This selective filtering significantly reduces the search space, leading to enhanced matching accuracy and efficiency in the overall process.
In experiments, we use VeriFinger \cite{verifinger} as a representative keypoint based matcher, which provides high quality minutiae extraction and matching functions.
Samples with pose differences greater than an optimal threshold (determined through exhaustive parameter search) will be excluded in advance.
In other words, the comparison scores of these detected abnormal situations are set to infinitely small.

On the other hand, we also assessed the impact of pose rectification on recognition schemes using fixed-length representations.
The current SOTA method FDD \cite{pan2024fixed} is utilized on behalf of these approaches.
In the recognition process, the input image is first rigidly transformed based on corresponding estimated pose. 
Subsequently, one-dimensional representation vectors are extracted and the matching similarity between pairs is calculated.

\begin{table*}[!t]
	\renewcommand\arraystretch{1.3}
	\belowrulesep=-0.2pt
	\aboverulesep=-0.2pt
	\caption{
    Verification performance (\%) on DPF (\fbox{partial fingerprints} with different rotation ranges) using different fingerprint poses.
    The four groups, from top to bottom, represent default configuration and methods that use only ridge patches, only capacitive images, and a combination of both.}
	\label{tab:verification-partial-dpf}
	\vspace{-0.4cm}
	\begin{center}
		\begin{threeparttable}
            \setlength{\tabcolsep}{5pt}
            \begin{tabular}{ c| c |ccc|ccc|ccc|ccc}
            \toprule
            \multirow{2}{*}[-.0mm]{\textbf{Matcher}} 
            & \multirow{2}{*}[-.0mm]{\textbf{Method}} 
            & \multicolumn{3}{c|}{{\textbf{[$\mathbf{-45^\circ,45^\circ}$]}}}
            & \multicolumn{3}{c|}{{\textbf{[$\mathbf{-90^\circ,90^\circ}$]}}}
            & \multicolumn{3}{c|}{{\textbf{[$\mathbf{-135^\circ,135^\circ}$]}}}
            & \multicolumn{3}{c}{{\textbf{[$\mathbf{-180^\circ,180^\circ}$]}}}
            \\
            \cmidrule(lr){3-5}\cmidrule(lr){6-8}\cmidrule(lr){9-11}\cmidrule(lr){12-14}
            {} & {}
            & \scriptsize\textbf{EER} & \scriptsize\textbf{FNMR\tnote{\,1}} & \scriptsize\textbf{FNMR\tnote{\,2}}
            & \scriptsize\textbf{EER} & \scriptsize\textbf{FNMR\tnote{\,1}} & \scriptsize\textbf{FNMR\tnote{\,2}}
            & \scriptsize\textbf{EER} & \scriptsize\textbf{FNMR\tnote{\,1}} & \scriptsize\textbf{FNMR\tnote{\,2}}
            & \scriptsize\textbf{EER} & \scriptsize\textbf{FNMR\tnote{\,1}} & \scriptsize\textbf{FNMR\tnote{\,2}}
            \\
            \midrule
            \multirow{6}{*}[-.0mm]{VeriFinger \cite{verifinger}} 
            & No Registration
            & 2.24 & 4.39 & 13.94
            & 2.06 & 3.88 & 14.81
            & 2.24 & 4.74 & 15.14
            & 2.32 & 5.19 & 17.17
            \\
            \cmidrule{2-14}
            {} & JointNet \cite{yin2021joint}
            & 2.13 & 3.84 & 13.08
            & 2.07 & 3.70 & 14.19
            & 2.60 & 4.81 & 15.74
            & 2.87 & 5.71 & 18.08
            \\
            {} & GridNet \cite{duan2023estimating}
            & 2.06 & 3.42 & 13.12
            & 2.02 & 3.59 & 14.10
            & 2.78 & 5.23 & 16.08
            & 3.71 & 7.01 & 19.19
            \\
            \cmidrule{2-14}
            {} & Cap-MLP \cite{xiao2015estimating}
            & 2.24 & 3.80 & 13.44 
            & 2.35 & 4.50 & 15.37
            & 6.56 & 10.82 & 20.68
            & 6.31 & 10.21 & 20.93
            \\
            {} & Cap-CNN \cite{mayer2017estimating,he2024TrackPose}
            & 2.04 & 3.40 & 13.11
            & 2.12 & 4.05 & 14.97
            & 4.80 & 8.64 & 18.74
            & 4.56 & 8.04 & 20.16
            \\
            \cmidrule[0.6pt]{2-14}
            {} & \cellcolor{black!10}{Proposed}
            & \cellcolor{black!10}\textbf{1.27} & \cellcolor{black!10}\textbf{1.61} & \cellcolor{black!10}\textbf{9.50}
            & \cellcolor{black!10}\textbf{1.11} & \cellcolor{black!10}\textbf{1.30} & \cellcolor{black!10}\textbf{8.41}
            & \cellcolor{black!10}\textbf{1.13} & \cellcolor{black!10}\textbf{1.26} & \cellcolor{black!10}\textbf{8.92}
            & \cellcolor{black!10}\textbf{0.96} & \cellcolor{black!10}\textbf{0.92} & \cellcolor{black!10}\textbf{9.09}
            \\
            \hline
            \hline
            \multirow{6}{*}[-.0mm]{FDD \cite{pan2024fixed}}
            & No Registration
            & 25.43 & 67.77 & 85.32
            & 36.67 & 84.20 & 93.53
            & 44.30 & 90.29 & 96.03
            & 44.05 & 92.49 & 97.46
            \\
            \cmidrule{2-14}
            {} & JointNet \cite{yin2021joint}
            & 20.72 & 53.97 & 74.40
            & 17.42 & 42.98 & 62.33
            & 23.74 & 56.50 & 74.82
            & 28.90 & 67.27 & 83.11
            \\
            {} & GridNet \cite{duan2023estimating}
            & 14.34 & 38.54 & 58.24
            & 19.04 & 47.84 & 66.17
            & 31.50 & 69.64 & 82.82
            & 40.46 & 85.99 & 93.74
            \\
            \cmidrule{2-14}
            {} & Cap-MLP \cite{xiao2015estimating}
            & 23.82 & 58.48 & 76.88
            & 36.05 & 82.12 & 91.37
            & 44.62 & 88.56 & 95.10
            & 44.15 & 90.60 & 96.58
            \\
            {} & Cap-CNN \cite{mayer2017estimating,he2024TrackPose}
            & 23.66 & 59.14 & 78.27
            & 27.12 & 61.97 & 79.13
            & 43.33 & 88.71 & 95.31
            & 44.14 & 92.18 & 97.34
            \\
            \cmidrule[0.6pt]{2-14}
            {} & \cellcolor{black!10}{Proposed}
            & \cellcolor{black!10}\textbf{5.70} & \cellcolor{black!10}\textbf{16.71} & \cellcolor{black!10}\textbf{37.41}
            & \cellcolor{black!10}\textbf{6.40} & \cellcolor{black!10}\textbf{19.48} & \cellcolor{black!10}\textbf{40.47}
            & \cellcolor{black!10}\textbf{6.75} & \cellcolor{black!10}\textbf{20.39} & \cellcolor{black!10}\textbf{44.14}
            & \cellcolor{black!10}\textbf{6.87} & \cellcolor{black!10}\textbf{21.37} & \cellcolor{black!10}\textbf{46.42}
            \\
            \bottomrule
            \end{tabular}
        \begin{tablenotes}
            \item[]$\phantom{x}^{1}$ FNMR@FMR=1e-3, $\phantom{x}^{2}$ FNMR@FMR=1e-4.
        \end{tablenotes}
		\end{threeparttable}
	\end{center}
\end{table*}

\begin{table}[!t]
	\renewcommand\arraystretch{1.3}
	\belowrulesep=-0.2pt
	\aboverulesep=-0.2pt
	\caption{
    Verification performance (\%) on \fbox{partial fingerprints} from PCF.
    The grouping rules are the same as Table \ref{tab:verification-partial-dpf}.}
	\label{tab:verification-partial-pcf}
	\vspace{-0.4cm}
	\begin{center}
		\begin{threeparttable}
            \setlength{\tabcolsep}{6pt}
            \begin{tabular}{ c| c |ccc}
            \toprule
            \textbf{Matcher} & \textbf{Method}
            & \scriptsize\textbf{EER} & \scriptsize\textbf{FNMR\tnote{\,1}} & \scriptsize\textbf{FNMR\tnote{\,2}}
            \\
            \midrule
            \multirow{6}{*}[-.0mm]{VeriFinger \cite{verifinger}} 
            & No Registration
            & 5.69 & 17.51 & 33.40
            \\
            \cmidrule{2-5}
            {} & JointNet \cite{yin2021joint}
            & 5.90 & 16.53 &  33.65
            \\
            {} & GridNet \cite{duan2023estimating}
            & 7.21 & 18.77 &  35.31
            \\
            \cmidrule{2-5}
            {} & Cap-MLP \cite{xiao2015estimating}
            & 8.18 & 20.33 & 36.75 
            \\
            {} & Cap-CNN \cite{mayer2017estimating,he2024TrackPose}
            & 7.85 & 19.72 & 36.19
            \\
            \cmidrule[0.6pt]{2-5}
            {} & \cellcolor{black!10}{Proposed}
            & \cellcolor{black!10}\textbf{4.09} & \cellcolor{black!10}\textbf{12.07} & \cellcolor{black!10}\textbf{30.08}
            \\
            \hline
            \hline
            \multirow{6}{*}[-.0mm]{FDD \cite{pan2024fixed}}
            & No Registration
            & 45.62 & 89.55 & 95.04
            \\
            \cmidrule{2-5}
            {} & JointNet \cite{yin2021joint}
            & 34.08 & 82.27 & 92.66
            \\
            {} & GridNet \cite{duan2023estimating}
            & 41.27 & 89.65 & 95.50
            \\
            \cmidrule{2-5}
            {} & Cap-MLP \cite{xiao2015estimating}
            & 44.01 & 91.53 & 96.80
            \\
            {} & Cap-CNN \cite{mayer2017estimating,he2024TrackPose}
            & 43.81 & 91.31 & 96.15
            \\
            \cmidrule[0.6pt]{2-5}
            {} & \cellcolor{black!10}{Proposed}
            & \cellcolor{black!10}\textbf{17.92} & \cellcolor{black!10}\textbf{56.01} & \cellcolor{black!10}\textbf{76.54}
            \\
            \bottomrule
            \end{tabular}
        \begin{tablenotes}
            \item[]$\phantom{x}^{1}$ FNMR@FMR=1e-3, $\phantom{x}^{2}$ FNMR@FMR=1e-4.
        \end{tablenotes}
		\end{threeparttable}
	\end{center}
\end{table}

\subsubsection{Evaluation on Fingerprint Verification}

In line with Section \ref{subsec:pose-estimation-performance}, we evaluate the performance of pose estimation methods on both plain and partial fingerprints.
This enables us to thoroughly evaluate their effectiveness across different situations.
The results at different rotation ranges on \emph{DPF} are reported in Table \ref{tab:verification-plain-dpf} and \ref{tab:verification-partial-dpf}, respectively.
Additionally, Table \ref{tab:verification-partial-pcf} presents a further comparison on \emph{PCF}.
When the pose difference is small, \emph{GridNet} \cite{duan2023estimating} exhibits certain advantages.
As the rotation angle increases, the performance of \emph{JointNet} \cite{yin2021joint} becomes more stable.
In more challenging scenarios involving partial fingerprints, previous works, whether based on ridge patches or capacitive images, have demonstrated unsatisfactory performance.
At the same time, our proposed DRACO showcases impressive and comprehensive advancements.
We attribute it to the decoupled pose representation and collaborative dual-modal guidance developed in this paper.
It is worth mentioning that fixed-length representation based matcher \cite{pan2024fixed} still has a gap in partial fingerprints compared to minutiae based solutions \cite{verifinger}.
Nevertheless, the introduction of our pose estimation method shows a significant relative improvement and shows attractive potential for future refinements.

\subsubsection{Evaluation on Fingerprint Indexing}

\begin{figure*}[!t]
	\centering
	\subfloat[]{\includegraphics[width=.23\linewidth]{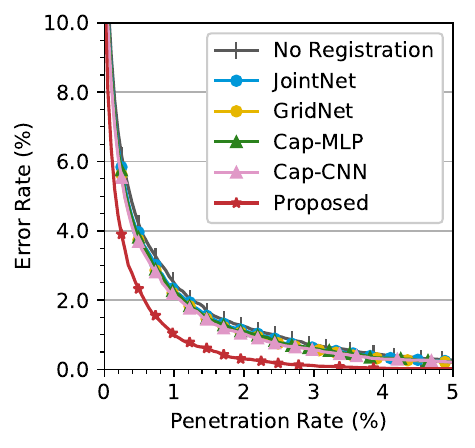}%
	\vspace{-1.5mm}}
	\hfil
	\subfloat[]{\includegraphics[width=.23\linewidth]{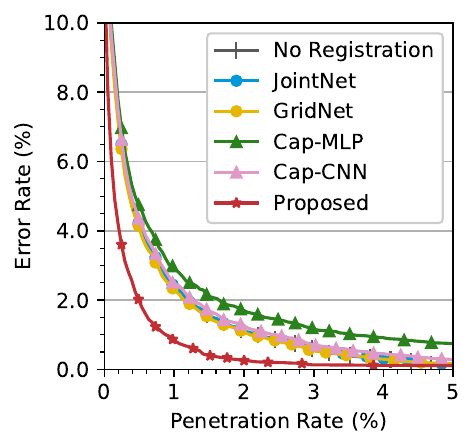}%
	\vspace{-1.5mm}}
    \hfil
    \subfloat[]{\includegraphics[width=.23\linewidth]{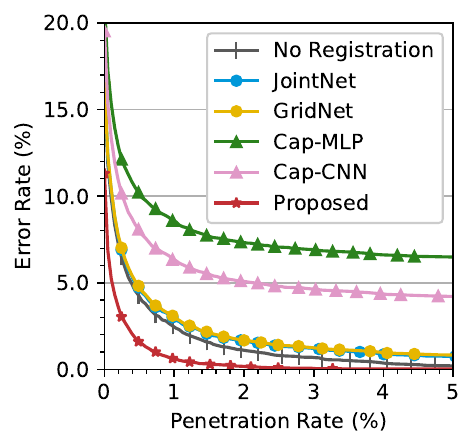}%
	\vspace{-1.5mm}}
	\hfil
	\subfloat[]{\includegraphics[width=.23\linewidth]{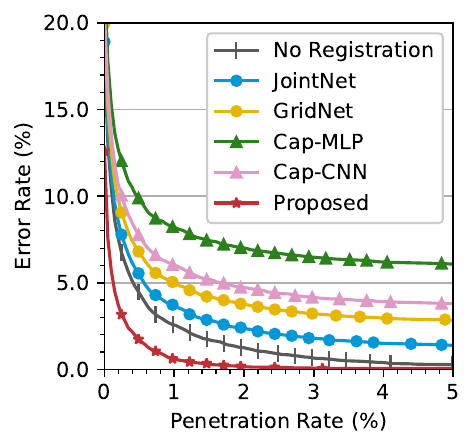}%
	\vspace{-1.5mm}}

	\caption{The VeriFinger \cite{verifinger} based fingerprint indexing performance with corresponding pose constraint on \fbox{partial fingerprints} from \emph{DPF} under different rotation ranges: (a) $[45^\circ,45^\circ]$, (b) $[90^\circ,90^\circ]$, (c) $[135^\circ,135^\circ]$, (d) $[180^\circ,180^\circ]$. 
    Different input modalities are distinguished by the shape of markers.}
	\label{fig:indexing-partial-dpf}
\end{figure*}

\begin{figure}[!t]
    \centering
    \includegraphics[width=0.95\linewidth]{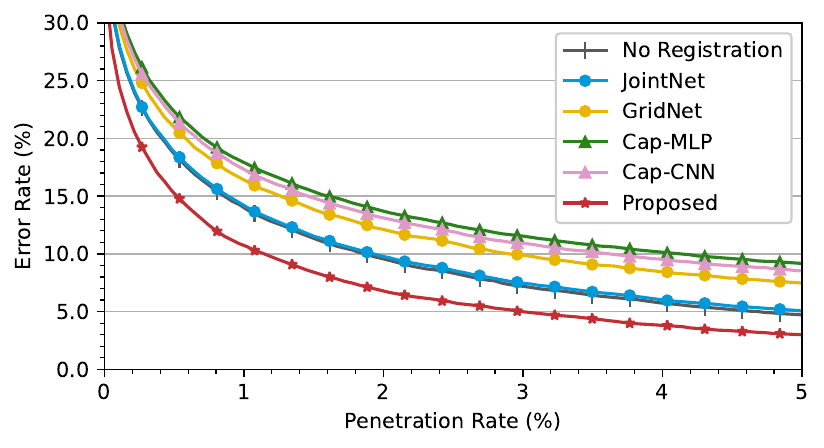}
    \caption{The VeriFinger \cite{verifinger} based fingerprint indexing performance with corresponding pose constraint on  \fbox{partial fingerprints} from  \emph{PCF}. 
    Different input modalities are distinguished by the shape of markers.
    }
    \label{fig:indexing-partial-pcf}
\end{figure}

Similarly, we further examine the role of fingerprint pose estimation algorithms in the indexing system.
Experiments are conducted solely on the minutiae based matcher, as this approach demonstrates superior performance and is the most commonly used in related works \cite{maltoni2022handbook}.
Fig. \ref{fig:indexing-plain-dpf} and \ref{fig:indexing-partial-dpf} illustrate the indexing performance for full fingerprint and partial fingerprint scenarios on \emph{DPF}, respectively, across different rotation ranges. 
The results on \emph{PCF} are shown in Fig. \ref{fig:indexing-partial-pcf}.
Comparisons in these curves indicate that appropriate pose estimation can effectively improve the indexing accuracy. 
In scenarios with a large rotation range or restricted effective areas, previous works face greater challenges and even have negative impacts in some cases.
Notably, our method excels as the most stable and accurate across all scenarios, which shows sufficient positive effects in all tests.

\subsection{Ablation Study}

\subsubsection{Pose Representation Form}

\begin{table}[!t]
	\renewcommand\arraystretch{1.3}
	\belowrulesep=-0.2pt
	\aboverulesep=-0.2pt
	\caption{
    Ablation study ($^\circ$) of pose representation form on DPF (\fbox{full fingerprints} with rotation range of $[-180^\circ,180^\circ]$).
    }
	\label{tab:ablation-representation}
	\vspace{-0.4cm}
	\begin{center}
		\begin{threeparttable}
            \setlength{\tabcolsep}{14pt}
            \begin{tabular}{ c | c  c  c c}
            \toprule
            \multirow{2}{*}[-.0mm]{\textbf{Loss\scriptsize{\textbackslash}Head}}
            & \multicolumn{2}{c}{\textbf{Regressor}}
            & \multicolumn{2}{c}{\textbf{Classifier}}
            \\
            \cmidrule(lr){2-3} \cmidrule(lr){4-5}
            {}
            & \textbf{ang}
            & \textbf{tan}
            & \textbf{max}
            & \textbf{sum}
            \\
            \midrule
            MSE (ang)
            & 13.2 & 7.9 & - & 7.3 \\
            MSE (tan)
            & 12.3 & 7.1 & - & 3.3 \\
            JS
            & \scriptsize\textbackslash & \scriptsize\textbackslash & 2.2 & 2.0 \\
            CE
            & \scriptsize\textbackslash & \scriptsize\textbackslash & 2.0 & \textbf{1.8} \\

            \bottomrule
            \end{tabular}
        \begin{tablenotes}
            \item[] `\scriptsize\textbackslash' indicates that corresponding item is not applicable.
            \item[] `-' indicates that corresponding process does not converge.
        \end{tablenotes}
		\end{threeparttable}
	\end{center}
\end{table}

We begin by thoroughly investigating the impact of pose representation and supervision forms on plain fingerprints of \emph{DPF}.
Table \ref{tab:ablation-representation} presents the ablation study for rotation as a representative example.
For the regression task head, two specific expressions are compared: (1) \emph{ang}: directly predicting the angle, and (2) \emph{tan}: predicting sine and cosine values and indirectly calculating the angle through the arctangent function.
Similarly, two types of mean squared error (MSE) are used as available losses.
The results indicate that representing angles using trigonometric functions yields improved performance, which is consistent with the explanation in Section \ref{subsec:disentagled-pose-representation}.
On this basis, we further transform the task into probability estimation and explore two classification heads for prediction: (1) \emph{max}: the highest response across all categories is directly selected as the result, and (2) \emph{sum}: the quantified probability distribution and class embedding are weighted and summed as Equation \ref{eq:distribution2value}.
Two classic distribution metrics, cross entropy (CE) and Jensen-Shannon divergence (JS), are supplemented as candidate supervisors.
The comparison strongly demonstrates the superiority of our proposed pose representation of probability distribution forms and verifies the rationality of the analysis in Section \ref{subsec:disentagled-pose-representation}.

\subsubsection{Feature Extraction \& Fusion}
Furthermore, the mechanisms and modules proposed in this paper for modal fusion and feature extraction enhancement are verified in Table \ref{tab:ablation-fusion}.
In the first group of experiments, the significant complementarity between ridge patches and capacitive images is reconfirmed.
On this basis, we compare different fusion strategies in the second group.
Three typical fusion schemes are evaluated, including
(1) \emph{E. MoE}: treating each expert as equally important, with their results directly summed, 
(2) \emph{F. MoE}: establishing a set of learnable parameters as globally fixed weights assigned to the experts,
(3) \emph{A. MoE}: dynamically assigning sample-specific adaptive weights to different experts through the router depicted in Fig. \ref{fig:network}.
Strategy (3) outperforms the others, which is understandable as it clearly exhibits the highest  flexibility and generalization ability.

The aim of the last comparison group is to assess the effectiveness of knowledge transfer concept introduced in Section \ref{subsec:knowledge-transfer}.
The supervision scheme based on features and responses \cite{yang2023categories} is additionally examined.
From the third and fourth lines to the bottom, it can be seen that directly approximating numerical values from the feature space proved ineffective, likely due to the significant domain difference between full and partial fingerprints.
The response based approach even has adverse effects. 
A convincing explanation is that it disrupts the probability distribution introduced by Equation \ref{eq:loss-pose}, which may lead to inaccurate representations.
Finally, knowledge transfer based on comparative relationships highlights the structured connections between samples, effectively incorporating higher-level semantic information and enhancing model performance.

\begin{table}[!t]
	\renewcommand\arraystretch{1.3}
	\belowrulesep=-0.2pt
	\aboverulesep=-0.2pt
	\caption{
    Ablation study of feature extraction and fusion on DPF (\fbox{partial fingerprints} with rotation range of $[-180^\circ,180^\circ]$).
    The gray background indicates the optimal strategy combination within each group, which serves as the basis for the following groups.
    }
	\label{tab:ablation-fusion}
	\vspace{-0.4cm}
	\begin{center}
		\begin{threeparttable}
            \setlength{\tabcolsep}{5pt}
            \begin{tabular}{ cc  c  c | cc}
            \toprule
            \multicolumn{2}{c}{\textbf{Used Modal}}
            & \multirow{2}{*}[-.0mm]{\makecell[c]{\textbf{Fusion}\\ \textbf{Strategy}\tnote{\:\dag}}} 
            & \multirow{2}{*}[-.0mm]{\makecell[c]{\textbf{Knowledge}\\ \textbf{Transfer}}}
            & \multicolumn{2}{c}{\textbf{Avg. Error}} 
            \\
            \cmidrule(lr){1-2} \cmidrule(lr){5-6}
            \textbf{Ridge} & \textbf{Cap} 
            & {} 
            & {}
            & \textbf{trans (px)} & \textbf{rot ($^\circ$)}
            \\
            \midrule
            \checkmark & {}
            & \scriptsize\textbackslash & \scriptsize\textbackslash
            & 25.4 & 14.1
            \\
            {} & \checkmark
            & \scriptsize\textbackslash & \scriptsize\textbackslash
            & 75.7 & 68.1
            \\
            \cellcolor{black!10}{\checkmark} & \cellcolor{black!10}{\checkmark}
            & \cellcolor{black!10}{\scriptsize\textbackslash} & \cellcolor{black!10}{\scriptsize\textbackslash}
            & \cellcolor{black!10}{23.5} & \cellcolor{black!10}{8.7}
            \\
            \cmidrule{1-6}
            \checkmark & \checkmark
            & \scriptsize{E. MoE} & \scriptsize\textbackslash
            & 22.4 & 7.7
            \\
            \checkmark & \checkmark
            & \scriptsize{F. MoE} & \scriptsize\textbackslash
            & 22.9 & 7.3
            \\
            \cellcolor{black!10}{\checkmark} & \cellcolor{black!10}{\checkmark}
            & \cellcolor{black!10}{\scriptsize{A. MoE}} & \cellcolor{black!10}{\scriptsize\textbackslash}
            & \cellcolor{black!10}{21.3} & \cellcolor{black!10}{6.8}
            \\
            \cmidrule{1-6}
            \checkmark & \checkmark
            & \scriptsize{A. MoE} & \scriptsize{Feature}
            & 21.0 & 6.5
            \\
            \checkmark & \checkmark
            & \scriptsize{A. MoE} & \scriptsize{Response}
            & 21.9 & 6.8
            \\
            \cellcolor{black!10}{\checkmark} & \cellcolor{black!10}{\checkmark}
            & \cellcolor{black!10}{\scriptsize{A. MoE}} & \cellcolor{black!10}{\scriptsize{Relation}}
            & \cellcolor{black!10}{\textbf{19.8}} & \cellcolor{black!10}{\textbf{5.5}}
            \\
            \bottomrule
            \end{tabular}
        \begin{tablenotes}
            \item[\dag] Abbreviations `E.', `F.' and `A.' respectively represent Equal, Fixed-Weight and Adaptive.
        \end{tablenotes}
		\end{threeparttable}
	\end{center}
\end{table}

\subsection{Efficiency}
Model size and inference speed of different fingerprint pose estimation algorithms on \emph{PCF} are listed in Table \ref{tab:efficiency}.
The time covers a complete process from inputting a sample to outputting the corresponding pose information, which is measured on a single NVIDIA GeForce RTX 3090 GPU by setting the batch size to 1, with an Intel Xeon E5-2680 v4 CPU @ 2.4 GHz. 
All algorithms are implemented in Python (Pytorch). 
It can be seen that our method exhibits comparable efficiency while delivering high estimation performance, thereby highlighting its attractive practical value.

\begin{table}[!t]
	\renewcommand\arraystretch{1.3}
	\belowrulesep=-0.2pt
	\aboverulesep=-0.2pt
	\caption{
    Model size and average time cost of different fingerprint pose estimation algorithms when processing PCF.
    Methods in each group use different modal input.
    }
	\label{tab:efficiency}
	\vspace{-0.4cm}
	\begin{center}
		\begin{threeparttable}
            \setlength{\tabcolsep}{12pt}
            \begin{tabular}{ c | c c }
            \toprule
            \textbf{Method}
            & \textbf{Param (M)}
            & \textbf{Times (ms)}
            \\
            \midrule
            JointNet \cite{yin2021joint}
            & 5.30 & 19.7
            \\
            GridNet \cite{duan2023estimating}
            & 14.2 & 16.8
            \\
            \cmidrule{1-3}
            CNN-MLP \cite{xiao2015estimating}
            & 0.17 & 3.6
            \\
            CNN-Cap \cite{mayer2017estimating,he2024TrackPose}
            & 1.70 & 4.4
            \\
            \cmidrule[0.6pt]{1-3}
            \cellcolor{black!10}{Proposed}
            & \cellcolor{black!10}{9.65} & \cellcolor{black!10}{26.1}
            \\
            \midrule
            
            \bottomrule
            \end{tabular}
		\end{threeparttable}
	\end{center}
\end{table}

\section{Conclusion}
In this paper, we propose DRACO, a novel partial fingerprint pose estimation method under dual-modal collaborative guidance of ridge patches and capacitive images, which are captured by under-screen fingerprint sensor and touch sensors of smartphones.
Unlike previous single modal based approaches, we demonstrate the strong complementarity between these two modalities and present an effective framework to integrate and leverage their combined strengths.
Specifically, relationship based knowledge transfer and MoE strategies are employed to enhance the network's feature extraction and fusion capabilities.
Furthermore, we reformulate fingerprint pose representation as a decoupled probability distribution, significantly improving prediction accuracy.
Extensive experiments on multiple databases show that DRACO surpasses state-of-the-art methods in both precision and robustness.
Future work will investigate deeper integration of fingerprint pose estimation with other related tasks and downstream processes, particularly improving its synergy with feature extraction and matching algorithms.


%

{
	\bibliographystyle{IEEEtran}
	\bibliography{egbib}{}
}

\clearpage

\onecolumn

\renewcommand\appendixname{Supplementary Materials}
\setcounter{page}{1}

\appendix

\renewcommand\thefigure{\Alph{section}.\arabic{figure}}
\setcounter{figure}{0}

\renewcommand\thetable{\Alph{section}.\arabic{table}}
\setcounter{table}{0}

\begin{table*}[h]
	\renewcommand\arraystretch{1.3}
	\belowrulesep=-0.2pt
	\aboverulesep=-0.2pt
	\caption{
    Alignment error under different fingerprint poses on DPF (\fbox{full fingerprints} with different rotation ranges).
    Suffixes `fp' indicate that only the corresponding branch is used.}
	\label{tab:pose-plain}
	\vspace{-0.4cm}
	\begin{center}
		\begin{threeparttable}
				\setlength{\tabcolsep}{12pt}
                \begin{tabular}{c | c c | c c | c c | c c}
                \toprule
                \multirow{2}{*}[-.0mm]{\textbf{Method}} 
                & \multicolumn{2}{c|}{{\textbf{[$\mathbf{-45^\circ,45^\circ}$]}}}
                & \multicolumn{2}{c|}{{\textbf{[$\mathbf{-90^\circ,90^\circ}$]}}}
                & \multicolumn{2}{c|}{{\textbf{[$\mathbf{-135^\circ,135^\circ}$]}}}
                & \multicolumn{2}{c}{{\textbf{[$\mathbf{-180^\circ,180^\circ}$]}}}
                \\
                \cmidrule(lr){2-3}\cmidrule(lr){4-5}\cmidrule(lr){6-7}\cmidrule(lr){8-9}
                {}
                & \scriptsize\textbf{trans (px)} & \scriptsize\textbf{rot ($^\circ$)}
                & \scriptsize\textbf{trans (px)} & \scriptsize\textbf{rot ($^\circ$)}
                & \scriptsize\textbf{trans (px)} & \scriptsize\textbf{rot ($^\circ$)}
                & \scriptsize\textbf{trans (px)} & \scriptsize\textbf{rot ($^\circ$)}
                \\
                \midrule
                No Registration
                & 83.4 & 22.7
                & 85.5 & 45.5
                & 85.1 & 67.4
                & 88.3 & 90.8
                \\
                Faster-RCNN \cite{ouyang2017fingerprint}
                & 218.4 & 22.3
                & 224.4 & 27.0
                & 118.3 & 66.6
                & 221.5 & 100.1
                \\
                STN \cite{engelsma2021learning,grosz2024AFRNet}
                & 90.1 & 16.2
                & 85.9 & 41.8
                & 85.7 & 67.6
                & 89.2 & 92.0
                \\
                JointNet \cite{yin2021joint}
                & 16.3 & 4.6
                & 18.8 & 7.5
                & 23.0 & 13.5
                & 22.8 & 19.9
                \\
                GridNet \cite{duan2023estimating}
                & 14.9 & 4.2
                & 19.7 & 6.2
                & 70.9 & 28.6
                & 83.7 & 62.3
                \\

                \cellcolor{black!10}{Proposed-fp}
                & \cellcolor{black!10}\textbf{14.5} & \cellcolor{black!10}\textbf{1.8}
                & \cellcolor{black!10}\textbf{15.0} & \cellcolor{black!10}\textbf{2.0}
                & \cellcolor{black!10}\textbf{14.1} & \cellcolor{black!10}\textbf{1.8}
                & \cellcolor{black!10}\textbf{14.2} & \cellcolor{black!10}\textbf{1.8}
                \\
                
                \bottomrule
                \end{tabular}
		\end{threeparttable}
	\end{center}
\end{table*}

\begin{table*}[h]
	\renewcommand\arraystretch{1.3}
	\belowrulesep=-0.2pt
	\aboverulesep=-0.2pt
	\caption{
    Verification performance (\%) on DPF (\fbox{full fingerprints} with different rotation ranges) using different fingerprint poses.
    Suffixes `fp' indicate that only the corresponding branch is used.}
	\label{tab:verification-plain-dpf}
	\vspace{-0.4cm}
	\begin{center}
		\begin{threeparttable}
            \setlength{\tabcolsep}{5pt}
            \begin{tabular}{ c| c |ccc|ccc|ccc|ccc}
            \toprule
            \multirow{2}{*}[-.0mm]{\textbf{Matcher}} 
            & \multirow{2}{*}[-.0mm]{\textbf{Method}} 
            & \multicolumn{3}{c|}{{\textbf{[$\mathbf{-45^\circ,45^\circ}$]}}}
            & \multicolumn{3}{c|}{{\textbf{[$\mathbf{-90^\circ,90^\circ}$]}}}
            & \multicolumn{3}{c|}{{\textbf{[$\mathbf{-135^\circ,135^\circ}$]}}}
            & \multicolumn{3}{c}{{\textbf{[$\mathbf{-180^\circ,180^\circ}$]}}}
            \\
            \cmidrule(lr){3-5}\cmidrule(lr){6-8}\cmidrule(lr){9-11}\cmidrule(lr){12-14}
            {} & {}
            & \scriptsize\textbf{EER} & \scriptsize\textbf{FNMR\tnote{\,1}} & \scriptsize\textbf{FNMR\tnote{\,2}}
            & \scriptsize\textbf{EER} & \scriptsize\textbf{FNMR\tnote{\,1}} & \scriptsize\textbf{FNMR\tnote{\,2}}
            & \scriptsize\textbf{EER} & \scriptsize\textbf{FNMR\tnote{\,1}} & \scriptsize\textbf{FNMR\tnote{\,2}}
            & \scriptsize\textbf{EER} & \scriptsize\textbf{FNMR\tnote{\,1}} & \scriptsize\textbf{FNMR\tnote{\,2}}
            \\
            \midrule
            \multirow{4}{*}[-.0mm]{VeriFinger\cite{verifinger}} 
            & No Registration
            & 0.71 & 0.42 & 1.90
            & 0.71 & 0.55 & 1.95
            & 0.64 & 0.34 & 1.84
            & 0.66 & 0.43 & 1.84
            \\
            {} & JointNet \cite{yin2021joint}
            & 0.64 & 0.32 & 1.90
            & 0.64 & 0.32 & 1.95
            & 0.60 & 0.23 & \textbf{1.66}
            & 0.85 & 0.79 & 2.20
            \\
            {} & GridNet \cite{duan2023estimating}
            & 0.63 & 0.32 & 1.90
            & 0.63 & 0.32 & 1.95
            & 0.70 & 0.48 & 1.98
            & 4.07 & 4.43 & 5.81
            \\
            {} & \cellcolor{black!10}{Proposed-fp}
            & \cellcolor{black!10}\textbf{0.59} & \cellcolor{black!10}\textbf{0.26} & \cellcolor{black!10}\textbf{1.82}
            & \cellcolor{black!10}\textbf{0.61} & \cellcolor{black!10}\textbf{0.24} & \cellcolor{black!10}\textbf{1.76}
            & \cellcolor{black!10}\textbf{0.54} & \cellcolor{black!10}\textbf{0.15} & \cellcolor{black!10}\textbf{1.66}
            & \cellcolor{black!10}\textbf{0.54} & \cellcolor{black!10}\textbf{0.13} & \cellcolor{black!10}\textbf{1.70}
            \\
            \hline
            \hline
            \multirow{4}{*}[-.0mm]{FDD\cite{pan2024fixed}} 
            & No Registration
            & 30.66 & 61.02 & 69.73
            & 40.14 & 83.44 & 89.03
            & 44.41 & 90.05 & 94.45
            & 44.95 & 92.80 & 96.40
            \\
            {} & JointNet \cite{yin2021joint}
            & 1.53 & 1.72 & 4.34
            & 2.67 & 3.79 & 7.06
            & 6.72 & 11.66 &19.05
            & 12.19 & 22.13 & 30.36
            \\
            {} & GridNet \cite{duan2023estimating}
            & 1.34 & \textbf{1.49} & 3.62
            & 2.57 & 3.48 & 5.84
            & 30.00 & 55.44 & 63.11
            & 41.40 & 85.19 & 90.73
            \\
            {} & \cellcolor{black!10}{Proposed-fp}
            & \cellcolor{black!10}\textbf{1.32} & \cellcolor{black!10}{1.50} & \cellcolor{black!10}\textbf{3.48}
            & \cellcolor{black!10}\textbf{1.19} & \cellcolor{black!10}\textbf{1.35} & \cellcolor{black!10}\textbf{3.67}
            & \cellcolor{black!10}\textbf{1.18} & \cellcolor{black!10}\textbf{1.29} & \cellcolor{black!10}\textbf{3.65}
            & \cellcolor{black!10}\textbf{1.07} & \cellcolor{black!10}\textbf{1.14} & \cellcolor{black!10}\textbf{3.34}
            \\
            \bottomrule
            \end{tabular}
        \begin{tablenotes}
            \item[]$\phantom{x}^{1}$ FNMR@FMR=1e-3, $\phantom{x}^{2}$ FNMR@FMR=1e-4.
        \end{tablenotes}
		\end{threeparttable}
	\end{center}
\end{table*}

\begin{figure*}[h]
	\centering
	\subfloat[]{\includegraphics[width=.23\linewidth]{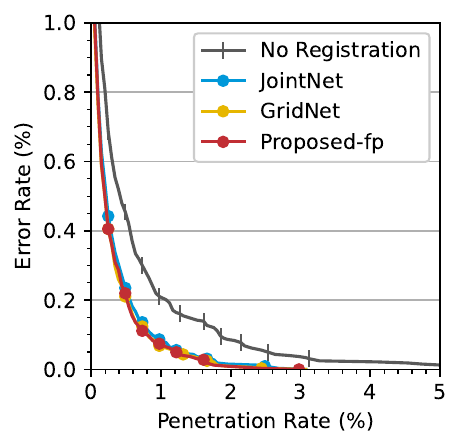}%
	\vspace{-1.5mm}}
	\hfil
	\subfloat[]{\includegraphics[width=.23\linewidth]{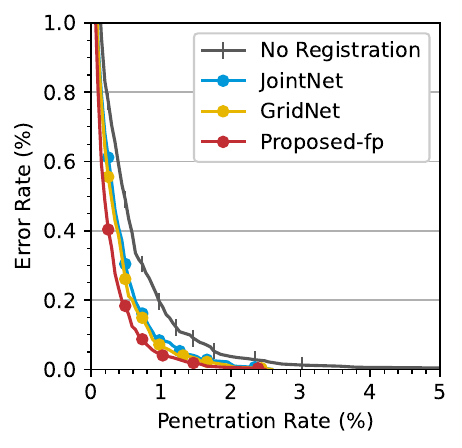}%
	\vspace{-1.5mm}}
    \hfil
    \subfloat[]{\includegraphics[width=.23\linewidth]{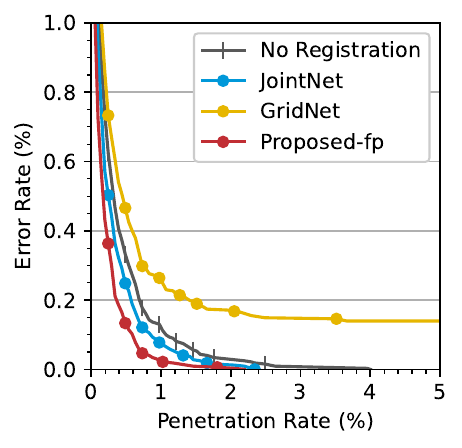}%
	\vspace{-1.5mm}}
	\hfil
	\subfloat[]{\includegraphics[width=.23\linewidth]{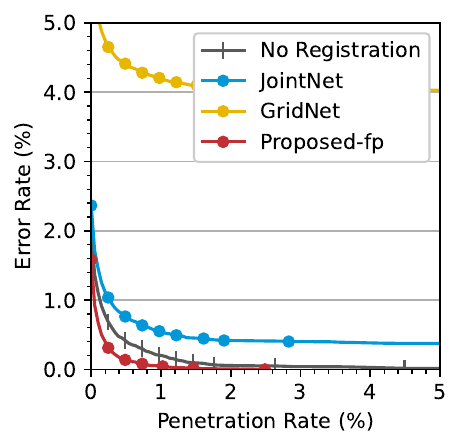}%
	\vspace{-1.5mm}}

	\caption{The VeriFinger \cite{verifinger} based fingerprint indexing performance with corresponding pose constraint on \fbox{full fingerprints} from \emph{DPF} under different rotation ranges: (a) $[45^\circ,45^\circ]$, (b) $[90^\circ,90^\circ]$, (c) $[135^\circ,135^\circ]$, (d) $[180^\circ,180^\circ]$.
    Different input modalities are distinguished by the shape of markers.}
	\label{fig:indexing-plain-dpf}
\end{figure*}

\end{document}